\documentclass[10pt]{wlscirep}
\usepackage[utf8]{inputenc}
\usepackage[T1]{fontenc}
\usepackage{xr}
\usepackage{commath,graphicx,afterpage,caption,subfigure,amsmath,amsfonts,amssymb,algorithmic,mathtools}
\usepackage{mathrsfs,amsthm,cite,multirow,bm,multicol,booktabs,threeparttable,extpfeil}
\usepackage{etoolbox}
\usepackage{lineno}
\usepackage[most]{tcolorbox}
\usepackage[authoryear,round]{natbib}

\makeatletter

\newcommand*\patchAmsMathEnvForLineno[1]{
  \expandafter\pretocmd\csname #1\endcsname{\linenomath}{}{}
  \expandafter\apptocmd\csname end#1\endcsname{\endlinenomath}{}{}
}
\newcommand*\patchBothAmsMathEnvsForLineno[1]{
  \patchAmsMathEnvForLineno{#1}\patchAmsMathEnvForLineno{#1*}
}
\newcommand{\mykeywords}[1]{%
  \vspace{0.5em}
  \noindent\textbf{Keywords} #1
  \vspace{0.5em}
}
\AtBeginDocument{
  \patchBothAmsMathEnvsForLineno{align}
  \patchBothAmsMathEnvsForLineno{alignat}
  \patchBothAmsMathEnvsForLineno{gather}
  \patchBothAmsMathEnvsForLineno{multline}
  \patchBothAmsMathEnvsForLineno{flalign}
  \patchBothAmsMathEnvsForLineno{equation}
}
\makeatother

\theoremstyle{definition}

\title{Synthetic Data Generation for Brain-Computer Interfaces: Overview, Benchmarking, and Future Directions}

\author[1]{Ziwei Wang}
\author[1]{Zhentao He}
\author[1]{Xingyi He}
\author[1]{Hongbin Wang}
\author[1]{Tianwang Jia}
\author[1]{Jingwei Luo}
\author[1]{Siyang Li}
\author[1,2]{Xiaoqing Chen}
\author[1,2,*]{Dongrui Wu}
\affil[1]{School of Artificial Intelligence and Automation, Huazhong University of Science and Technology, Wuhan, China}
\affil[2]{Zhongguancun Academy, Beijing, China}
\affil[*]{Corresponding Author: Dongrui Wu (drwu09@gmail.com)}

\begin{abstract}
\textbf{Deep learning has achieved transformative performance across diverse domains, largely driven by large-scale and high-quality training data. In contrast, the development of brain-computer interfaces (BCIs) is fundamentally constrained by limited, heterogeneous, and privacy-sensitive neural recordings. Generating synthetic yet physiologically plausible brain signals has therefore emerged as a promising strategy to mitigate data scarcity, improve model generalization, and support data-efficient BCIs. This survey provides a comprehensive review of synthetic brain data generation for BCIs, covering methodological taxonomies, benchmark experiments, evaluation metrics, key applications, and future directions. We systematically categorize existing generation approaches into four types: signal-transformation-based, feature-based, model-based, and translation-based generation, and discuss their characteristics, advantages, and limitations. Furthermore, we benchmark representative brain signal generation approaches across four BCI paradigms, including motor imagery, epileptic seizure detection, steady-state visually evoked potentials, and auditory attention detection, to provide an objective comparison of their downstream utility. We also summarize evaluation principles for generated brain signals from multiple perspectives, including signal realism, physiological plausibility, downstream utility, and privacy preservation. Finally, we discuss the potential and challenges of current generation approaches and outline future research directions toward accurate, data-efficient, generalizable, and privacy-aware BCI systems. The benchmark codebase is available at \url{https://github.com/wzwvv/DG4BCI}.
}
\end{abstract}

\begin{document}

\flushbottom
\maketitle
\mykeywords{Brain-computer interfaces $\cdot$ Synthetic data generation $\cdot$ Generative models $\cdot$ Electroencephalography $\cdot$ Data augmentation $\cdot$ Neural decoding}
\thispagestyle{empty}

\section{Introduction}
Brain-computer interface (BCI) serves as a direct communication pathway between a user's brain and an external device, enabling mapping, assisting, augmenting, and potentially restoring human cognitive and/or sensorymotor functions \citet{Ienca2018NB}. Despite rapid progress, real-world BCI systems still suffer from the limited data availability, strong individual variability, and non-stationarity of brain signals. The reliability of decoding models is closely tied to the quantity and quality of the available brain data.

\begin{figure}[htbp]\centering
\includegraphics[width=.5\linewidth,clip]{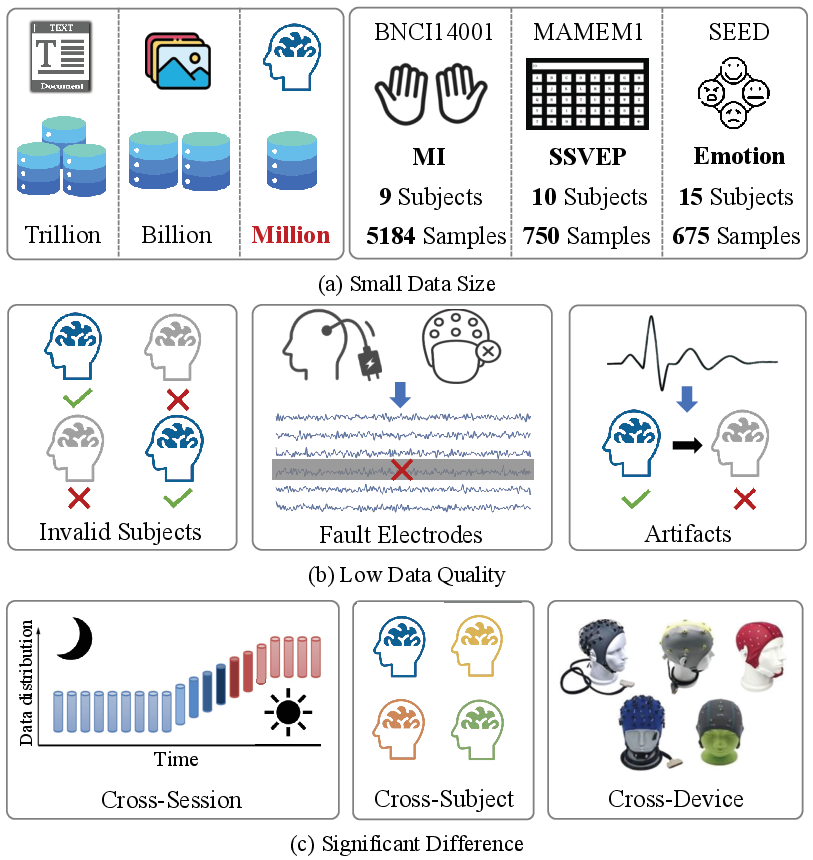}
\caption{Data scarcity issue in BCIs, owing to the small data size, low signal quality, and/or significant differences across sessions, subjects, and devices.} \label{fig:intro_data_scarcity}
\end{figure}

Recent success of deep learning underscores the importance of large, high-quality training data. For example, Google's trillion-word corpus has significantly boosted performance in language models \cite{Brown2020}. However, acquiring sufficient brain signals presents challenges, illustrated in Figure \ref{fig:intro_data_scarcity}.
\begin{itemize}
\item \textit{High signal acquisition costs}. For example, devices for collecting brain signals, especially intracranial signals, are expensive.
\item \textit{Difficulty in long-term collection}. Only a limited amount of data can be acquired in each session due to the discomfort caused by the acquisition devices during long-term collection.
\item \textit{Low signal quality}. Raw brain signals are often non-stationary and prone to noise. Artifacts such as electrooculographic signals from eye movements \cite{wang2015removal,shahbakhti2021vme}, electromyographic activity from facial muscles, cardiac artifacts, baseline drift, and electromagnetic interference frequently contaminate the signals \cite{tatum2011artifact,chen2025self}. These factors degrade signal quality and may impair downstream decoding. In addition, obtaining precise labels is challenging because it is difficult to ensure that subjects consistently perform the intended tasks.
\item \textit{Significant signal heterogeneity}. Unlike image or text data, brain signals vary significantly across subjects, datasets, and devices, limiting model generalization.
\item \textit{Privacy concerns}. Legal and regulatory restrictions may limit the sharing of sensitive brain data, as transferring it across institutions can risk exposing private information.
\end{itemize}

The above limitations hinder the development of robust, generalizable decoding models for BCIs \cite{luo2020data}. The scarcity and heterogeneity of brain data, together with privacy barriers, motivate the development of brain signal generation techniques capable of synthesizing informative and diverse samples to support learning under limited supervision.

Based on the electrode placement, brain signals can be categorized into non-invasive, partially invasive, and invasive ones. Non-invasive brain signals are considered the safest, without any surgical procedures. Partially invasive brain signals involve the implantation of electrodes beneath the scalp, without penetrating the brain tissue. Invasive BCIs require surgical implantation of electrodes into the brain, offering the highest resolution and precision for neural signal recording. Each acquisition modality exhibits distinct spatial and temporal characteristics, as summarized in Table~\ref{tab:signals}. Among them, electroencephalography (EEG) remains the most widely adopted owing to its safety, affordability, and portability. Magnetoencephalography (MEG) and functional near-infrared spectroscopy (fNIRS) complement EEGs with enhanced spatial or hemodynamic information, while electrocorticography (ECoG) and stereoelectroencephalography (SEEG) provide higher fidelity but at a high surgical cost.
\begin{itemize}
\item EEG is one of the most widely used non-invasive techniques due to its cost-effectiveness and ease of application. It captures the brain's electrical activity through scalp electrodes.
\item MEG measures neuromagnetic activity with millisecond temporal precision and moderate spatial resolution. It enables non-invasive mapping of cortical dynamics critical for studying real-time neural computation and oscillatory processes.
\item fNIRS measures blood oxygenation changes via near-infrared light and offers moderate spatial and temporal resolution. It is portable and less susceptible to electromagnetic interference, making it suitable for both clinical and field-based research settings.
\item ECoG is collected from electrodes placed on the exposed surface of the brain, providing high spatial resolution and direct access to cortical areas. It offers better signal quality than EEG, but carries some risks of surgical intervention.
\item SEEG is collected from electrodes implanted through small burr holes into deep brain regions, primarily offering detailed monitoring of neural activity in patients with brain diseases.
\end{itemize}

\begin{table}[htpb]  \center \setlength{\tabcolsep}{1mm}
\small
\caption{Characteristics of brain signals in BCIs.}  \label{tab:signals}
\begin{tabular}{c|c|c|c|c|c|c}
\toprule
Electrode Placement & Signal Type & Temporal Resolution & Spatial Resolution & Cost & Universality & Setup Duration\\
\midrule
\multirow{3}{*}{Non-invasive} & EEG & High & Low & Cheap & High & Quick (Dry) / Moderate (Wet)\\
& fNIRS & Moderate & Moderate & Moderate & Moderate & Quick\\
& MEG & High & Moderate & Expensive & Low & Moderate\\
\midrule
Partially Invasive & ECoG & High & High & Expensive & Low & Short-term implant\\
\midrule
Invasive & SEEG & High & High & Expensive & Low & Short-term implant\\
\bottomrule
\end{tabular}
\end{table}

This survey aims to provide a unified view of synthetic data generation to improve BCI model training. Specifically, we organize existing studies into four categories: signal-transformation-based, feature-based, model-based, and translation-based approaches, and discuss how different generation strategies expand training data from signal, feature, model, and modality perspectives. Beyond taxonomy, we further summarize evaluation metrics, benchmark representative approaches, and discuss key applications in data-efficient, generalizable, and privacy-aware BCIs. The benchmark comparison is conducted on four representative paradigms: motor imagery (MI), an active BCI modality relevant to neurorehabilitation; epileptic seizure detection (ESD), a medical-grade application involving spontaneous pathological signals; steady-state visually evoked potentials (SSVEP), a frequency-coded visual stimulation paradigm characterized by periodic neural responses for target identification; and auditory attention detection (AAD), a selective-listening paradigm for decoding users' attended speech streams. Rather than claiming to fully characterize all dimensions of generation quality, our benchmark focuses on the downstream utility of generated data for BCI decoding, while broader issues like physiological plausibility, distributional realism, and privacy preservation are discussed as critical evaluation dimensions. Overall, this survey seeks to clarify the types of synthetic data generation for BCIs, when they are effective, how they should be evaluated, and the challenges that remain for future research.

\begin{figure}[htbp]\centering
\includegraphics[width=\linewidth,clip]{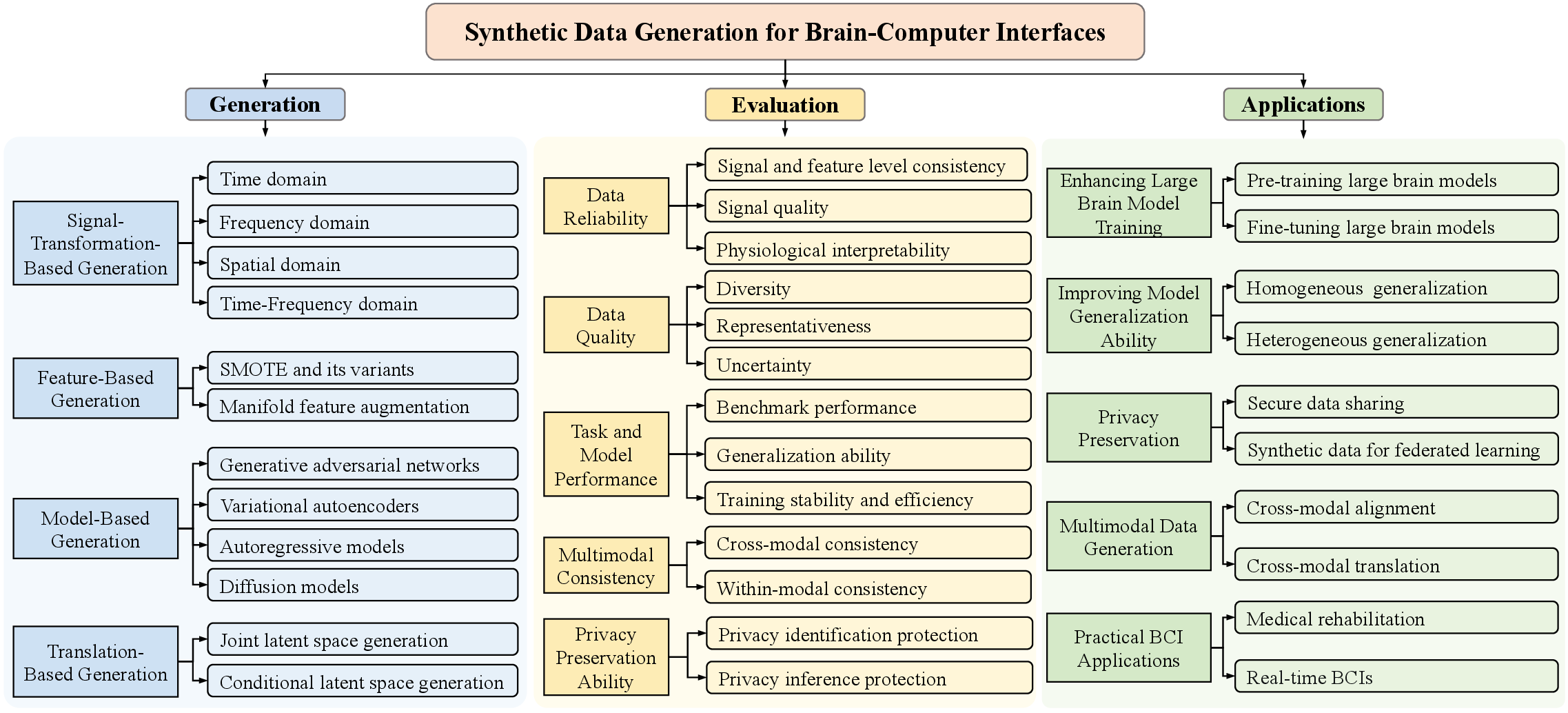}
\caption{Overall organization of this survey.} \label{fig:paper_framework}
\end{figure}

The remainder of this paper is organized as follows: Section~\ref{sect:sdg} reviews data generation algorithms in BCIs. Section~\ref{sect:benchmark} benchmarks existing brain signal generation algorithms. Section~\ref{sect:sde} summarizes evaluation metrics of the generated data. Section~\ref{sect:app} presents the representative applications. Section~\ref{sect:discuss} summarizes the insights of existing work and presents some possible future directions. Section~\ref{sect:conclusions} draws conclusions. The overall organization of this survey is illustrated in Figure~\ref{fig:paper_framework}.

\section{Synthetic Data Generation for BCIs}\label{sect:sdg}

\begin{figure*}[htbp]\centering
\includegraphics[width=.9\linewidth,clip]{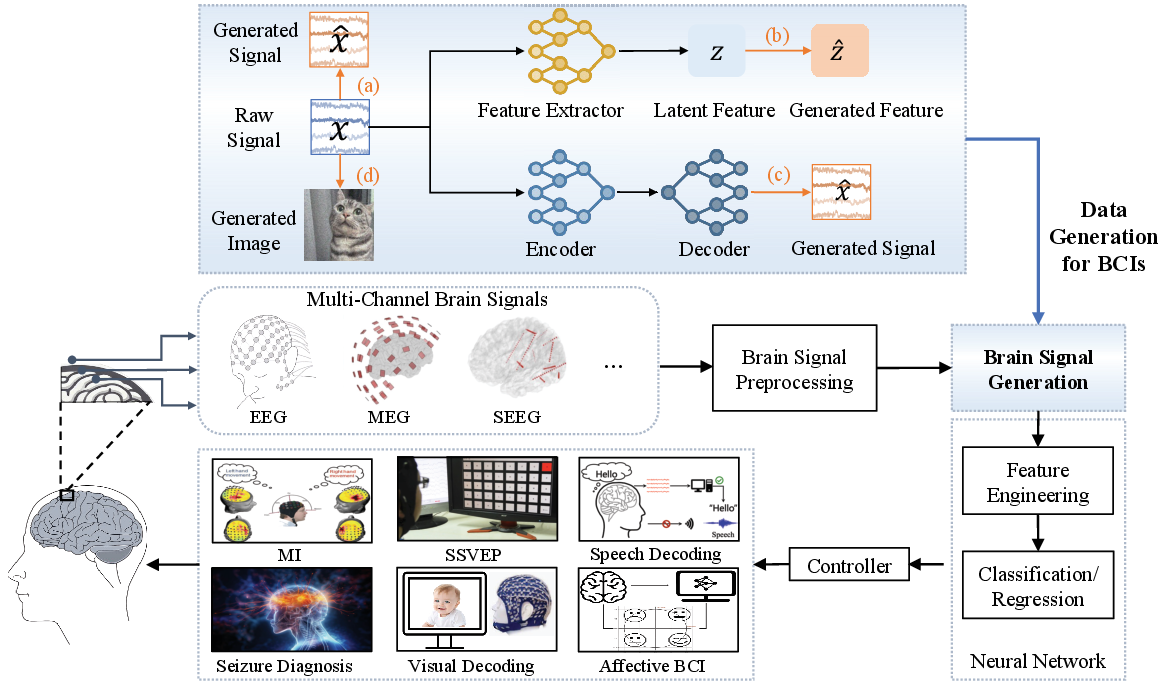}
\caption{Data generation driven machine learning pipeline for BCIs, which includes brain signal acquisition, data preprocessing, data generation, feature engineering, and classification/regression. The latter two components can be unified into a single end-to-end neural network. Data generation approaches are categorized into four types: (a) signal-transformation-based generation, (b) feature-based generation, (c) model-based generation, and (d) translation-based generation.} \label{fig:intro}
\end{figure*}

\begin{figure}[htbp]\centering
\includegraphics[width=.8\linewidth,clip]{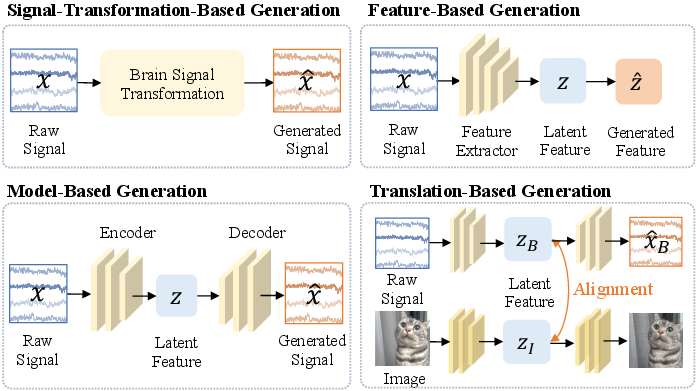}
\caption{Four types of data generation approaches for brain signals.} \label{fig:four_types_gene}
\end{figure}

The data generation driven machine learning pipeline for BCIs is depicted in Figure~\ref{fig:intro}, where the generation algorithms can be categorized into four types: signal-transformation-based, feature-based, model-based, and translation-based approaches. Details of the four types are further illustrated in Figure~\ref{fig:four_types_gene}.

\subsection{Signal-Transformation-Based Generation}

Signal-transformation-based brain signal generation produces synthetic signals by applying predefined transformations to existing brain signals. These transformations are usually designed in the temporal, spectral, spatial, or time-frequency domains and can be implemented without explicitly training a generative model. From a broad machine learning perspective, such augmentation-derived samples can be regarded as synthetic training data, as they expand the effective data distribution available for model learning.

Within this category, some approaches are general-purpose perturbations, such as noise injection, amplitude scaling, and voltage inversion, which mainly improve model robustness through regularization. Other strategies are knowledge-informed transformations that explicitly incorporate neurophysiological or signal-processing priors. For example, spatial transformations may exploit hemispheric symmetry in MI \cite{Wang2024CR}, while time-frequency transformations can preserve rhythmic structures related to task-specific neural responses \cite{Wang2025CSDA}. Such priors help maintain the biological plausibility of generated signals while increasing data diversity.

Following Rommel \textit{et al.} \cite{rommel2022data}, EEG augmentation approaches can be broadly grouped into time, frequency, and spatial domain approaches. Since time-frequency decomposition techniques can jointly characterize temporal dynamics and spectral patterns of brain signals, we further extend this taxonomy by introducing time-frequency domain approaches:
\begin{itemize}
\item Time domain approaches, which directly manipulate the temporal characteristics of brain signals. For example, Wang \textit{et al.} \cite{Wang2018} introduced Gaussian white noise to the original signals, Mohsenvand \textit{et al.} \cite{Mohsenvand2020} randomly masked portions of EEG segments, and Zhang \textit{et al.} \cite{Zhang2022MSDT} applied minor scaling and voltage inversion.
\item Frequency domain approaches, which first transform brain signals into the frequency domain, perform augmentation, and then convert them back to the time domain. For example, Zhang \textit{et al.} \cite{Zhang2022MSDT} performed frequency shifting, Schwabedal \textit{et al.} \cite{Schwabedal2018} replaced the Fourier phases of EEG trials with random numbers, and Zhao \textit{et al.} \cite{zhao2022seizure} used the discrete cosine transform (DCT) to sample and recombine frequency bands.
\item Spatial domain approaches, which exploit the spatial structure of multi-channel brain signals. For example, Wang \textit{et al.} \cite{Wang2024CR} swapped symmetric hemisphere channels and their labels for left/right hand MI, Krell \textit{et al.} \cite{Krell2017} applied random interpolation to rotated channels, and Pei \textit{et al.} \cite{pei2021hs} recombined hemispheric channels.
\item Time-frequency domain approaches, which employ time-frequency decomposition techniques for data generation. For example, Wang \textit{et al.} \cite{Wang2025CSDA} generated brain signals with the discrete wavelet transform (DWT) and Hilbert-Huang transform (HHT). Both typically involve three steps: time-frequency decomposition, sub-signal reassembly, and temporal reconstruction. This direction remains underexplored and offers significant potential for future research.
\end{itemize}

\begin{figure}[htbp]\centering
\includegraphics[width=.6\linewidth,clip]{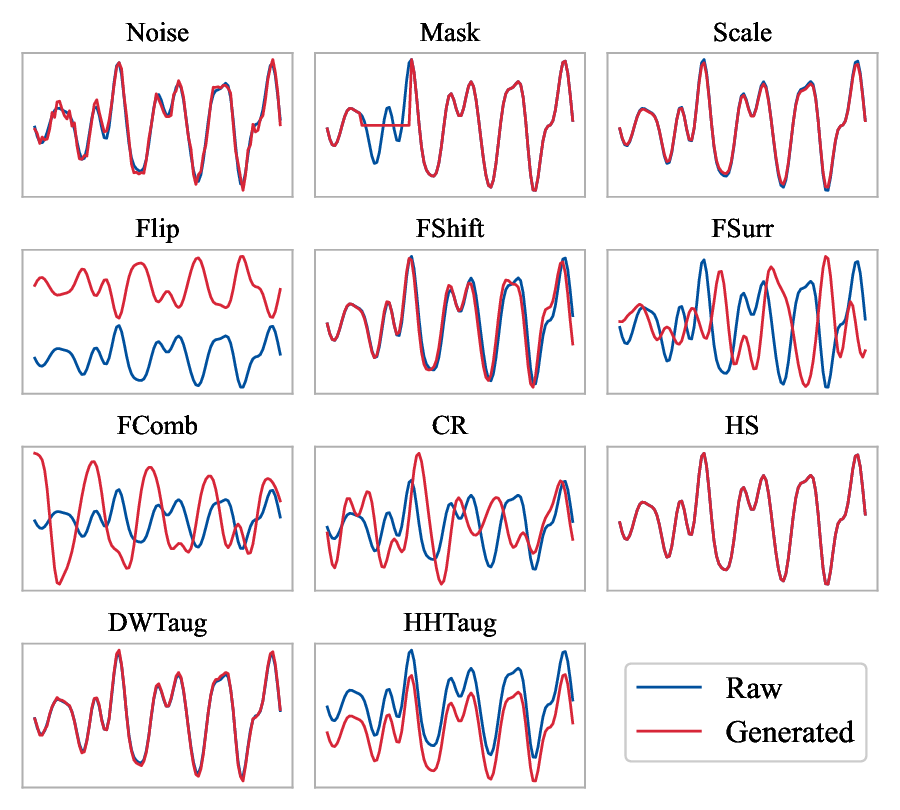}
\caption{Visualizations of brain signals before (blue lines) and after (red lines) eleven signal-transformation-based generation approaches, using 1-channel as an example.} \label{fig:kbDG}
\end{figure}

Signal-transformation-based generation approaches are summarized in Table~\ref{tab:kbDG}, and their temporal visualizations are shown in Figure~\ref{fig:kbDG}.

\begin{table*}[htpb]  \center
\setlength{\tabcolsep}{0.3mm}
\footnotesize
\caption{Four types of signal-transformation-based generation approaches.}  \label{tab:kbDG}
\begin{tabular}{l|l|l|l}
\toprule
Type & Approach & Description & Parameter\\
\midrule
\multirow{4}{*}{Time domain}
& Noise \cite{Wang2018} & Adding uniform noise to an EEG trial in the time domain & $C_{noise}$: the noise injection rate \\
& Mask \cite{Mohsenvand2020} & Masking a portion of EEG trials randomly & $C_{mask}$: the masking rate\\
& Scale \cite{Zhang2022MSDT} & Scaling the voltage of EEG trials with a minor coefficient & $C_{scale}$: the scaling rate\\
& Flip \cite{Zhang2022MSDT} & Performing voltage inversion & / \\
\midrule
\multirow{3}{*}{Frequency domain}
& FShift \cite{Zhang2022MSDT} & Shifting the frequency of EEG trials & $C_{fshift}$: the frequency scaling rate\\
& FSurr \cite{Schwabedal2018} & Replacing the Fourier phases of trials with random numbers & $C_{p}$: probability; $C_{m}$: noise magnitude\\
& FComb \cite{zhao2022seizure} & Applying discrete cosine transform and recombining bands & $C_{comb}$: the number of combining bands\\
\midrule
\multirow{2}{*}{Spatial domain}
& CR \cite{Wang2024CR} & Swapping symmetric left and right hemisphere channels & /\\
& HS \cite{Krell2017} & Recombining left and right channels from the same category & /\\
\midrule
\multirow{2}{*}{Time-Frequency domain}
& DWTaug \cite{Wang2025CSDA} & Decomposing trials by DWT and reassembling coefficients & $C_{level}$: the number of decomposition level \\
& HHTaug \cite{Wang2025CSDA} & Empirical mode decomposition and sub-signal reassembling & / \\
\bottomrule
\end{tabular}
\end{table*}

\subsection{Feature-Based Generation}

Feature-based generation approaches focus on generating synthetic features rather than raw signals. This strategy helps improve model generalization by enriching the feature space, playing a crucial role in enhancing the diversity of features.

The widely adopted techniques are the synthetic minority over-sampling technique (SMOTE) \cite{Wang2023TASA} and its variants \cite{Tseng2024}, which generate synthetic features by interpolating between existing minority-class samples. Feature-based generation approaches are particularly beneficial for seizure detection or anxiety state classification, where imbalanced datasets often degrade model performance. Borderline SMOTE and adaptive synthetic have been further explored for generating EEG features in mental stress detection \cite{Tseng2024}, ensuring that newly generated samples lie closer to the decision boundary, thereby improving classifier robustness.

Additionally, Mixup \cite{Zhang2018} extends the idea of SMOTE by applying linear interpolations to both features and labels, helping improve model generalization across varying conditions. Freer \textit{et al.} \cite{Freer2020} leveraged empirical mode decomposition to separate EEG signals into intrinsic modes for data generation. Manifold-based feature generation exploits the intrinsic geometric structure of the feature space by mapping features onto lower-dimensional manifolds and synthesizing new samples through distribution-aware sampling. Manifold Mixup \cite{Verma2019MMixup} enhances neural network training by performing linear interpolations within hidden representations, generating high-confidence synthetic features.

\subsection{Model-Based Generation}

Model-based generation is more flexible, commonly relying on probabilistic generative models to generate synthetic brain signals, based on the idea that the underlying distribution of brain signals can be modeled and sampled from sophisticated deep learning models.

\begin{figure}[htbp]\centering
\includegraphics[width=.6\linewidth,clip]{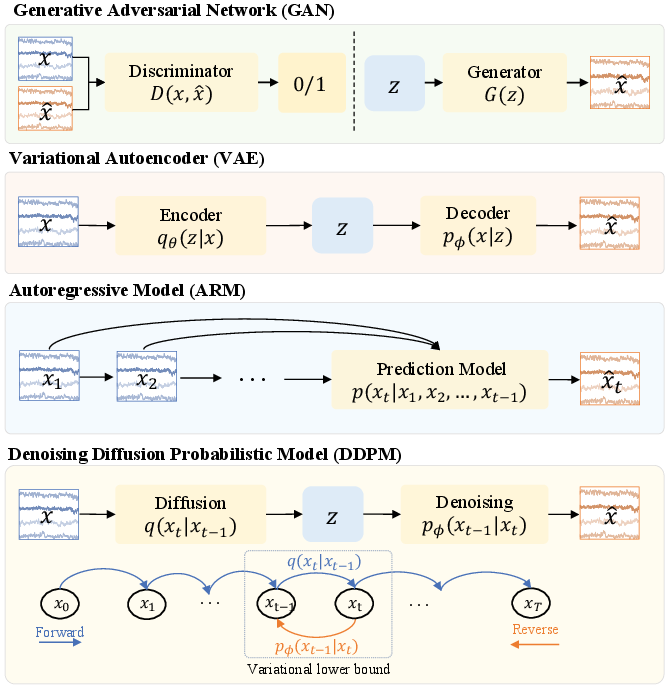}
\caption{Model-based generation approaches for brain signals, including GANS, VAEs, ARMs, and DDPMs.} \label{fig:mb_gene}
\end{figure}

The key advantage of model-based generation lies in its ability to model high-dimensional, nonlinear data distributions, such as those inherent in brain signals. These models go beyond basic data augmentation techniques by providing a deeper understanding of the signal's probabilistic structure. There are mainly four categories of approaches, including generative adversarial networks (GANs) \cite{ganin2016domain}, variational autoencoders (VAEs) \cite{kingma2014auto}, autoregressive models (ARMs) \cite{yang2019xlnet}, and denoising diffusion probabilistic models (DDPMs) \cite{ho2020denoising}, as illustrated in Figure~\ref{fig:mb_gene}. A comparison between existing model-based EEG generation studies is provided in Table~\ref{tab:mbDG}, including the model type, synthetic signal type, applied paradigm, dataset, implementation, and their gains.

\begin{table*}[htpb]  \center \setlength{\tabcolsep}{0.4mm}
\renewcommand{\arraystretch}{.95}
\footnotesize
\caption{Comparison of existing studies on brain signal generative models, including GANs, VAEs, ARMs, and DDPMs.}  \label{tab:mbDG}
\begin{tabular}{c|c|c|c|c|c|c}
\toprule
Model Type & Approach & Signal Type & Paradigm & Dataset & Setup & Improvement \\
\midrule
\multirow{25}{*}{GAN}
& cc-LSTM-GAN \cite{wen2023rapid} & Spike & Motor execution & Two Monkeys & Conditional & 16.0\% accuracy \\ 
& DCGAN \cite{fahimi2020generative} & EEG and ECG & Motor execution & Fourteen humans & Conditional & 7.3\% accuracy\\
& GDAL \cite{ko2022semi} & EEG & MI & III-3a, III-4a, and IV-2a & Unconditional & 2.0\% accuracy\\
& SSVEP-GAN\cite{figueiredo2023optimizing} & EEG & SSVEP & 32 humans & Conditional & 96s time-saving\\
& TOED-GAN \cite{zeng2024task} & EEG & SSVEP & Dial & Conditional & 18.5\% accuracy\\
& WGAN \cite{aznan2021leveraging} & EEG & SSVEP & NAO & Unconditional & 3.0\% accuracy\\
& \cite{ma2025brain} & EEG & Visual perception & CelebA & Conditional & 6.6\% accuracy\\
& \cite{lee2024enhanced} & EEG & Speech imagery & \cite{lee2020neural} & Unconditional & / \\
& NeuroTalk \cite{lee2023speech} & EEG & Speech imagery & \cite{lee2020neural} & Unconditional & reduced 0.6 RMSE \\
& CS-GAN \cite{song2021common} & EEG & MI & IV-2a & Unconditional & 15.9\% accuracy \\
& EEG-GAN \cite{hartmann2018eeg} & EEG & MI & \cite{hartmann2018eeg}& Unconditional & 0.08 inception score\\
& \cite{palazzo2017generative} & EEG & Image generation & \cite{palazzo2017generative} & Conditional & 40.5\% accuracy \\
& cWGAN \cite{luo2020data} & EEG & Emotion recognition & SEED, DEAP & Conditional & 4.9\% accuracy \\
& sWGAN \cite{luo2020data} & EEG & Emotion recognition & SEED, DEAP & Conditional & 7.4\% accuracy \\
& SleepEGAN \cite{cheng2024sleepegan} & EEG & Sleep stage classification & Sleep-EDF, SHHS & Unconditional & 7.4\% accuracy \\
\cmidrule{2-7}
& \multirow{4}{*}{\shortstack{SynSigGAN \\ \cite{hazra2020synsiggan}}} & EEG & Seizure prediction & Siena Scalp & Unconditional & / \\
& & ECG & Arrhythmia detection & MIT-BIH & Unconditional & / \\
& & EMG & Sleep stage classification  & Sleep-EDF & Unconditional & / \\
& & PPG & Respiratory rate estimation & BIDMC & Unconditional & / \\
\cmidrule{2-7}
& \multirow{2}{*}{\shortstack{WGAN-GP \\ \cite{venugopal2024boosting}}} & EEG & Mental state classification & MUSE & Unconditional & 6.5\% accuracy \\
& & ECG & Abnormality detection & Kaggle & Unconditional & 1.4\% accuracy \\
\cmidrule{2-7}
& BWGAN-GP \cite{xu2022bwgan} & EEG & RSVP & \cite{bird2018study} & Conditional & 3.7\% accuracy\\
& \cite{kavasidis2017brain2image} & EEG & Image generation & Private dataset  & Unconditional & / \\
& CESP \cite{rasheed2021generative} & EEG & Seizure prediction & CHB-MIT & Unconditional & 4.0\% sensitivity\\
\midrule
\multirow{10}{*}{VAE}
 & cVAE \cite{luo2020data} & EEG & Emotion recognition & SEED, DEAP & Conditional & 3.1\% accuracy \\
 & sVAE \cite{luo2020data} & EEG & Emotion recognition & SEED, DEAP & Conditional & 3.2\% accuracy \\
 & DEVAE-GAN \cite{tian2023dual} & EEG & Emotion recognition & SEED & Conditional & 5.0\% accuracy\\
 & EEG2Vec \cite{bethge2022eeg2vec} & EEG & Emotion recognition & SEED & Conditional & 3.0\% accuracy \\
 & CR-VAE \cite{li2023causal} & ECoG & Seizure detection & \cite{kramer2008emergent} & Conditional & reduced 0.014 RMSE \\
\cmidrule{2-7}
& \multirow{2}{*}{\shortstack{VAEEG \\ \cite{zhao2024vaeeg}}} & EEG & Seizure detection & TUSZ & Conditional & 1.4\% accuracy\\
& & EEG & Sleep stage classification & NCH & Conditional & 1.3\% accuracy\\
\cmidrule{2-7}
 & \cite{kavasidis2017brain2image} & EEG & Image generation & \cite{kavasidis2017brain2image}  & Unconditional & / \\
 & CVAE \cite{tibermacine2025conditional}& EEG & MI & OpenNeuro & Conditional & 1.1\% accuracy\\
 & VAE \cite{yildiz2022unsupervised} & iEEG & Seizure detection & Kaggle, TUH, CHB-MIT & Unconditional & 12.0\% accuracy \\
\midrule
\multirow{11}{*}{ARM}
& Neuro-GPT \cite{cui2024neuro} & EEG & MI & TUH & Unconditional & 9.3\% accuracy \\
\cmidrule{2-7}
& \multirow{2}{*}{\cite{bird2021synthetic}} & EEG & Mental state classification & \cite{bird2018study} & Conditional & 0.9\% accuracy \\
& & EMG & Gesture recognition & \cite{dolopikos2021electromyography} & Conditional & 0.3\% accuracy \\
\cmidrule{2-7}
& \cite{niu2021epileptic} & EEG & Seizure prediction & MIT & Conditional & 1.9\% accuracy\\
\cmidrule{2-7}
& \multirow{2}{*}{MEG-GPT \cite{huang2025meg}} & MEG & Cognitive BCI & Cam-CAN & Unconditional & 8.0\% accuracy \\
& & MEG & Visual perception & Wakeman-Henson & Unconditional & 5.0\% accuracy \\
\cmidrule{2-7}
& ChannelGPT2 \cite{csaky2024foundational} & MEG & Visual perception & \cite{cichy2016comparison} & Conditional & 3.3 \% accuracy \\
& Thought2Text \cite{mishra2025thought2text} & EEG, Text & Text generation & CVPR2017 & Conditional & 18.2 BLEU-1\\
& GET \cite{ali2024get} & EEG & MI & IV-2b & Unconditional & / \\
& Endemann \cite{endemann2022multivariate} & iEEG & Seizure detection & \cite{endemann2022multivariate} & Unconditional & / \\
\midrule
\multirow{17}{*}{DDPM}
& \cite{zhong2024enhanced} & EEG & MI & 64 humans & Unconditional & 3.2\% accuracy\\
& DDIM \cite{torma2025generative} & EEG & MI, SSVEP & IV-2a, VEPESS& Unconditional & 10.3\% accuracy\\
& CDASC-MI \cite{zhou2025generative} & EEG & MI & IV-2a, IV-2b & Unconditional & 6.0\% accuracy\\
&  \cite{dere2025motor} & EEG, EMG & MI & OpenBCI & Conditional & 3.4\% precision \\
& \cite{kim2024brain} & EEG & Speech imagery & \cite{lee2020neural} & Conditional & 15.3\% accuracy\\
& Diff-E \cite{kim2023diff} & EEG & Speech imagery & \cite{lee2020neural} & Conditional & 14.4\% accuracy \\
& DiffEEG \cite{shu2024data} & EEG & Seizure prediction & CHB-MIT & Conditional & 8.6\% AUC \\
\cmidrule{2-7}
& \multirow{2}{*}{AE-KL \cite{aristimunha2023synthetic}} & EEG & Sleep stage classification & SleepEDFx & Unconditional & reduced 11.6 FID \\
& & EEG & Sleep stage classification & SHHS & Unconditional & reduced 0.8 FID \\
\cmidrule{2-7}
& EEG-DIF \cite{jiang2025diffusion} & EEG & Seizure prediction & Physionet & Unconditional & 8.0\% accuracy \\
& \cite{klein2024synthesizing} & EEG & P300 & OpenBMI & Conditional & / \\
&  \cite{liu2025diffusion} & EEG & Driving behavior classification & Private dataset & Conditional & 7.1\% \\
& \cite{chetkin2024unconditional} & EEG & MI & IV-2a & Unconditional & / \\
& ControlNet \cite{postolache2025naturalistic} & EEG, Audio & Audio generation & NMED-T & Conditional & reduced 0.23 MSE \\
& Dreamdiffusion \cite{bai2023dreamdiffusion} & EEG, image & Image generation & ImageNet-EEG & Unconditional & 7.3\% \\
\cmidrule{2-7}
& \multirow{2}{*}{EEGDiffuser \cite{wang2026eegdiffuser}} & EEG & MI, Imagined speech decoding & IV-2a, BCIC2020-3 & Conditional & / \\
& & EEG & Emotion recognition & FACED & Conditional & 2.3\% accuracy \\
\bottomrule
\end{tabular}
\end{table*}

\subsubsection{Generative Adversarial Networks}

Introduced in \cite{goodfellow2014generative}, GANs have gained significant traction in the field of data generation due to their ability to produce high-fidelity samples, frequently used in areas like image generation \cite{mao2017least}, audio synthesis \cite{donahue2018adversarial}, time series augmentation \cite{yoon2019time}, and have been taken into consideration for EEG generation. They have proven effective in generating high-fidelity samples under proper training. However, their training process usually meets mode collapse and can be computationally expensive.

Fahimi \textit{et al.} \cite{fahimi2019towards} highlighted the effectiveness of GANs in EEG data generation, claiming that the generated EEG signals by GANs resemble the temporal, spectral, and spatial characteristics of real EEG \cite{fahimi2020generative}.
Wen \textit{et al.} \cite{wen2023rapid} proposed cc-LSTM-GAN to augment the monkey's spike data. EEG-GAN \cite{hartmann2018eeg} utilized GAN with a convolutional generator to synthesize EEG data, allowing the generation of raw EEG signals from random noise for MI classification. Cheng \textit{et al.} \cite{cheng2024sleepegan} proposed SleepEGAN to generate minority category EEG samples for sleep stage classification. Hazra and Byun \cite{hazra2020synsiggan} introduced SynSigGAN to generate four types of synthetic biomedical signals, including EEG signals for epileptic seizure classification. Arjovsky \textit{et al.} \cite{arjovsky2017wasserstein} proposed Wasserstein GAN (WGAN) to overcome model collapse. Gulrajani \textit{et al.} \cite{gulrajani2017improved} added the gradient penalty to WGAN (WGAN-GP) that could maintain more stable training. Palazzo \textit{et al.} \cite{palazzo2017generative} applied WGAN conditioned on EEG features for cross-modal image generation, using EEG as an auxiliary input. This approach allows the synthesis of images corresponding to specific brain states or stimuli, expanding the possibilities for visual decoding. Luo \textit{et al.} \cite{luo2020data} introduced two types of GANs, including conditional WGAN and selective WGAN, to augment EEG differential entropy features for emotion recognition. Venugopal and Faria \cite{venugopal2024boosting} validated the effectiveness of WGAN-GP in EEG and ECG generation for mental state classification. Xu \textit{et al.} \cite{xu2022bwgan} further verified that WGAN-GP could generate a balanced EEG dataset with improved performance for the rapid serial visual presentation classification task.

GANs consist of two modules: a generator $G$ to generate synthetic data and a discriminator $D$ to evaluate the authenticity of the generated data against real data. $G$ and $D$ are trained in an adversarial setting, allowing generator $G$ to gradually improve its ability to generate realistic data that can no longer be distinguished by the discriminator $D$. These opposing networks are simultaneously trained to maximize $log(D(x,\hat{x}))$ and minimize $log(1-D(G(z)))$, which can be formulated as a minimax problem:
\begin{align}
\min _{G} \max _{D}V(G, D)=E_{x \sim p_{x}} & {[\log (D(x,\hat{x}))]} \nonumber\\
&+E_{z \sim p_{z}}[\log (1-D(G(z)))],
\end{align}
where $E$ is the expectation operator, $\hat{x}$ the generated sample, $D(x, \hat{x})$ the probability of $\hat{x}$ belonging to the real $x$, $z$ a random noise input, and $\hat{x}=G(z)$. The cross-entropy loss is commonly used as the optimization objective function for the generator and discriminator, respectively formulated as:
\begin{align}
\mathcal{L}_G=-\log D(G(z)),
\end{align}
and
\begin{align}
\mathcal{L}_D=-\log D(x,\hat{x})-log(1-D(G(z))).
\end{align}

\subsubsection{Variational Autoencoders}

VAEs have also been applied to the generation of brain signals. Bethge \textit{et al.} \cite{bethge2022eeg2vec} introduced a conditional VAE that learns generative-discriminative representations from affective EEG data, enabling emotion-related signal generation. This framework can learn generative-discriminative representations from EEG signals and generate synthetic EEG signals that resemble real EEG data inputs, particularly for reconstructing low-frequency signal components. Li \textit{et al.} \cite{li2023causal} developed a causal recurrent VAE to construct a Granger causal graph from EEG signals, illustrating the ability of VAEs to not only generate brain signals but also capture complex data dependencies. This capability is crucial for understanding the causal relationships across brain regions.

Besides directly generating the raw signals, some works explored generating features in the latent space. Luo \textit{et al.} \cite{luo2020data} introduced two types of VAEs, conditional VAE and selective VAE, to augment EEG differential entropy features for emotion recognition. Tian \textit{et al.} \cite{tian2023dual} also proposed a dual encoder VAE-GAN  incorporating spatiotemporal features to augment differential entropy features for emotion recognition.

VAEs are latent variable generative models consisting of an encoder and a decoder. The encoder $q_{\boldsymbol{\theta}}(z|x)$ of VAE learns a lower-dimensional representation $z$ of the raw trial $x$, known as the latent space, with variational parameters $\boldsymbol{\theta}$. The decoder $p_{\boldsymbol{\phi}}(x|z)$ outputs the probability distribution of the raw trial based on the latent representation $z$, with its parameters $\boldsymbol{\phi}$. Deep latent variable models aim to maximize the probability of the raw trial $x$ by
\begin{align}
p(x)=\int p(x|z) p(z)dx.
\end{align}
However, the marginal probability of data under the model is typically intractable due to this integral not having an analytic solution or efficient estimator. In practice, $p(x|z)$ is nearly zero for most $z$ and contributes almost nothing to estimate $p(x)$. To overcome this, VAEs try to sample $z$ by approximating the posterior $p_{\boldsymbol{\phi}}(z|x)$ with $q_{\boldsymbol{\theta}}(z|x)$. $p_{\boldsymbol{\phi}}(x)$ can be defined as:
\begin{align}
&\log p_{\boldsymbol{\phi}}(x) \nonumber\\
 =&\mathbb{E}_{q_{\boldsymbol{\theta}}(z|x)}\left[\log p_{\boldsymbol{\phi}}(x)\right] 
 =\mathbb{E}_{q_{\boldsymbol{\theta}}(z|x)}\left[\log \left[\frac{p_{\boldsymbol{\phi}}(x,z)}{p_{\boldsymbol{\phi}}(z|x)}\right]\right] \nonumber\\
 =&\mathbb{E}_{q_{\boldsymbol{\theta}}(z|x)}\left[\log \left[\frac{p_{\boldsymbol{\phi}}(x,z)}{q_{\boldsymbol{\theta}}(z|x)}\right]\right] +\mathbb{E}_{q_{\boldsymbol{\theta}}(z|x)}\left[\log \left[\frac{q_{\boldsymbol{\theta}}(z|x)}{p_{\boldsymbol{\phi}}(z|x)}\right]\right],
\end{align}\label{eq:vae}
where the first term is the evidence lower bound (ELBO), and the second term is Kullback-Leibler (KL) divergence $D_{KL}\left(q_{\boldsymbol{\theta}}(z|x) \| p_{\boldsymbol{\phi}}(z|x)\right)$ used to measure the discrepancy between two distributions. Thus, it can be rewritten as
\begin{align}
\log p_{\boldsymbol{\phi}}(x) & =ELBO+D_{KL}\left(q_{\boldsymbol{\theta}}(z|x) \| p_{\boldsymbol{\phi}}(z|x)\right),
\end{align} \label{eq:vaes_short}
where the KL divergence is non-negative according to Jensen's inequality.

VAEs are well-suited for probabilistic modeling and generating smooth, continuous latent spaces. Their ability to model complex data distributions makes them ideal for generating brain signals conditioned on specific neural activities, such as emotion or cognitive tasks. Although VAEs can offer valuable insights into latent representations, they may struggle with generating highly detailed samples compared to GANs, especially when it comes to capturing fine-grained temporal or spatial features in brain signals.

\subsubsection{Autoregressive Models}

ARMs, such as generative pre-trained transformers (GPT) that are decoder-only transformers, are designed to generate data sequentially. ARMs handle long-range dependencies within brain signals, facilitating more accurate signal generation. Cui \textit{et al.} \cite{cui2024neuro} introduced a model comprising an EEG encoder and a GPT pre-trained on a large-scale dataset to reconstruct masked EEG trials, enabling realistic EEG generation by decoding temporal dependencies of EEG signals. Bird \textit{et al.} \cite{bird2021synthetic} employed multiple GPT-2 models to generate synthetic EEG signals, enhancing classification performance for unseen subjects. Similarly, Niu \textit{et al.} \cite{niu2021epileptic} used GPT to generate synthetic EEG signals for epileptic seizure prediction, aiding the early detection of abnormal brain activity. In ARMs, the model generates each trial conditioning on all previous ones:
\begin{align}
p(x)=\prod_{t=1}^{T} p\left(x_{i} \mid x_{1}, x_{2}, \ldots, x_{t-1};\boldsymbol{\theta}\right),
\end{align}
where $p(x_t|x_1,x_2,...,x_{t-1};\boldsymbol{\theta})$ is the probability of the current trial $x_t$ conditioned on all preceding trials, $\boldsymbol{\theta}$ the model parameters, and $T$ the total number of trials. The training objective of ARMs is to minimize the negative log-likelihood loss:
\begin{align}
\mathcal{L}_{ARM}=-\sum_{t=1}^{T} \log p\left(x_{t} \mid x_{1}, x_{2}, \ldots, x_{t-1} ;\boldsymbol{\theta}\right).
\end{align}

ARMs excel in capturing long-term temporal dependencies, making them suited for understanding the sequential patterns in brain signals. However, ARMs can be computationally intensive and often require large-scale training datasets to perform optimally.

\subsubsection{Denoising Diffusion Probabilistic Models}

DDPMs have emerged as a potential tool for identifying nuanced patterns within complex, high-dimensional EEG signals. Tosato \textit{et al.} \cite{tosato2023eeg} applied DDPMs to generate synthetic EEG data in both time and frequency domains, independent of channel count. This ability to handle EEG data across different domains enhances the flexibility and robustness of DDPMs for real-world BCIs. Further, DDPMs have been employed for cross-modal generation, where EEG signals are used to reconstruct high-quality images \cite{zeng2023dm}. Kim \textit{et al.} \cite{kim2024brain} utilized DDPMs in conjunction with the conditional autoencoder to analyze speech-related EEG signals.

DDPMs generate synthetic data by progressively denoising random noise in a sequence of steps, conditioning each step on previous iterations~\cite{ho2020denoising}. In DDPM, a sequence of noise coefficients $\beta_1, \beta_2, ..., \beta_T$ for Markov transition kernels is chosen, following patterns of constant, linear, or cosine schedules to generate high-quality samples. Following~\cite{ho2020denoising}, the forward diffusion steps are defined as:
\begin{align}
F_{t}\left(x_{t-1}, \beta_t\right)=q\left(x_{t} \mid x_{t-1}\right)=\mathcal{N}\left(x_{t} ; \sqrt{1-\beta_t} x_{t-1}, \beta_t \mathbf{I}\right),
\end{align}
where $F_t$ is the forward transition kernels at time $t$. With the composition of forward transition kernels form $x_0$ to $x_T$, the diffusion process adds Gaussian noises to brain signals by Markov kernel $q(x_t|x_{t-1})$:
\begin{align}
F\left(x_{0},\left\{\beta_{i}\right\}_{i=1}^{T}\right)=q\left(x_{1:T} \mid x_{0}\right)=\prod_{t=1}^{T} q\left(x_{t} \mid x_{t-1}\right).
\end{align}
The reverse denoising step, with learnable Gaussian kernels optimized by $\boldsymbol{\phi}$, is formulated as:
\begin{align}
R_{t}\left(x_{t}, \Sigma_{\boldsymbol{\phi}}\right)&=p_{\boldsymbol{\phi}}\left(x_{t-1} \mid x_{t}\right) \nonumber\\
&=\mathcal{N}\left(x_{t-1} ; \mu_{\boldsymbol{\phi}}\left(x_{t}, t\right), \Sigma_{\boldsymbol{\phi}}\left(x_{t}, t\right)\right),
\end{align}
where $R_t$ is the reverse transition kernels at time $t$, $\mu_{\boldsymbol{\phi}}$ and $\Sigma_{\boldsymbol{\phi}}$ are learnable mean and variance of the reverse Gaussian kernel, and $p_{\boldsymbol{\phi}}$ the reverse step distribution. Reverse steps from $x_T$ to $x_0$ is
\begin{align}
R\left(x_{T}, \Sigma_{\boldsymbol{\phi}}\right)=p_{\boldsymbol{\phi}}\left(x_{0: T}\right)=p\left(x_{T}\right) \prod_{t=1}^{T} p_{\boldsymbol{\phi}}\left(x_{t-1} \mid x_{t}\right).
\end{align}

DDPMs also require significant computational resources and have not yet been as widely adopted as GANs or VAEs for generating brain signals.

\subsection{Translation-Based Generation}

Translation-based generation involves synthesizing data by integrating information from additional modalities, typically through cross-modal generative models. In BCIs, this strategy is crucial for bridging diverse data types, such as brain signals \cite{du2020conditional}, text \cite{willett2021high}, speech \cite{defossez2023decoding}, images \cite{du2023decoding}, and other sensor data.

\begin{figure*}[htbp]\centering
\includegraphics[width=\linewidth,clip]{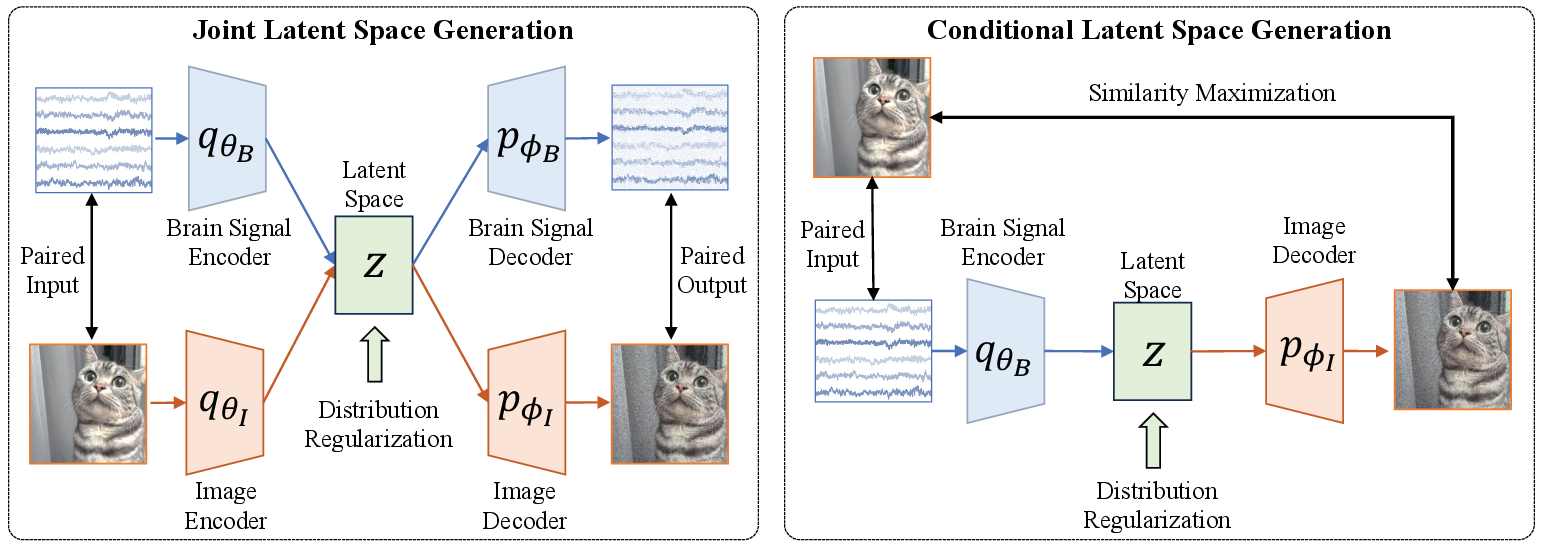}
\caption{Two types of translation-based generation for brain signals, taking two modalities of image and brain signal as an example.} \label{fig:translation_generation}
\end{figure*}

Generative models serve as the foundation for translation-based generation, learning the distributions across modalities. Typical approaches can be divided into the joint and conditional latent space generative models. Both leverage latent spaces for cross-modal data generation, but they differ in how to construct and utilize the latent space, as shown in Figure~\ref{fig:translation_generation}.

\subsubsection{Joint Latent Space Generation}

Joint latent space generation aims to map multiple modalities into a shared latent space, where data from each modality are encoded into a unified lower-dimensional representation. The shared latent space captures the relationships and dependencies between modalities. For example, in the brain-to-image paradigm, separate encoders $E_{\boldsymbol{\theta}_B}$ and $E_{\boldsymbol{\theta}_I}$ project brain signals and images into a common space $z$ with reduced dimensionality such that $E_{\boldsymbol{\theta}_B}(x^B), E_{\boldsymbol{\theta}_I}(x^I) \in \mathbb{R}^d$, where $x^B$ and $x^I$ are the brain signal and image pair inputs. The training process typically involves a Gaussian distribution for regularization:
\begin{align}
& \mathcal{L}_{\text{joint}} \left( (x^B, x^I); E_{\boldsymbol{\theta}_B}; D_{\smash{\boldsymbol{\phi}_B}}; E_{\boldsymbol{\theta}_I}; D_{\smash{\boldsymbol{\phi}_I}} \right) = \nonumber \\ & \hfill \mathbb{E}_{q_{\boldsymbol{\theta}_B}(z|x^B)} \left[ \log p_{\smash{\boldsymbol{\phi}_B}}(x^B|z) \right] + \mathbb{E}_{q_{\boldsymbol{\theta}_I}(z|x^I)} \left[ \log p_{\smash{\boldsymbol{\phi}_I}}(x^I|z) \right] \nonumber \\ & \hfill - D_{\text{KL}} \left( q_{\boldsymbol{\theta}_B}(z|x^B) \| p(z) \right) - D_{\text{KL}} \left( q_{\boldsymbol{\theta}_I}(z|x^I) \| p(z) \right),
\end{align}
where $q(z|x)$ is encoders' approximate posterior distributions, $p(x|z)$ decoders' prior likelihood distributions, $p(z)$ the prior distribution over the latent space assumed to be a standard Gaussian distribution, and $D_{\text{KL}}(q\|p)$ the KL divergence that regularizes the latent distribution.

\subsubsection{Conditional Latent Space Generation}

In contrast, conditional latent space generation focuses on generating the target modality based on a given input modality, without mapping both modalities into a shared latent space. This approach is useful when the target modality depends on specific conditions or context from the source modality. For example, an encoder $E_{\boldsymbol{\theta}_B}$ maps a brain signal input to a latent space $z$, and the image decoder $D_{\boldsymbol{\phi}_I}$ reconstructs the corresponding image. The training process is formulated as:
\begin{align}
& \mathcal{L}_{\text{conditional}} \left( (x^B, x^I); E_{\boldsymbol{\theta}_B}, D_{\smash{\boldsymbol{\phi}_I}} \right) \nonumber\\
= &\mathbb{E}_{q_{\boldsymbol{\theta}_B}(z | x^A)} \left[ \log p_{\smash{\boldsymbol{\phi}_I}}(x^I | z) \right] - D_{\text{KL}} \left( q_{\boldsymbol{\theta}_B}(z | x^B) \| p(z) \right).
\end{align}

Conditional models have shown significant promise in applications such as EEG signal reconstruction, brain-to-text generation, and multimodal image generation \cite{du2023decoding}. For instance, brain signals can be conditioned on a specific task to generate corresponding visual representations in brain-to-image tasks. In EEG signal reconstruction, a model conditioned on a specific cognitive state could generate synthetic EEG signals that closely match the brain's activity. In multimodal image generation, brain signals condition the model to generate images corresponding to specific mental images or visual stimuli. Du \textit{et al.} \cite{du2020conditional} proposed a brain-to-image translation framework that decodes features of brain data to reconstruct the perceived image.

Despite promising developments in cross-modal generation, several challenges remain. Training generative models with multiple modalities requires high-quality, well-aligned datasets, which are often hard to obtain, especially for brain signals. Additionally, models must address modal inconsistencies. In conclusion, translation-based generative models bridge brain signals with other modalities, facilitating effective multimodal interaction. However, achieving high-quality and contextually accurate cross-modal generation remains a key challenge.

\subsection{Comparison of Generation Approaches}

To summarize, Table~\ref{tab:compare_four} presents a comparison of the four brain signal generation approaches, providing an overview of their strengths, limitations, and key challenges.

\begin{table*}[htbp]
\centering \setlength{\tabcolsep}{0.7mm}
\caption{Comparison of four types of synthetic data generation approaches in BCIs.}
\small
\label{tab:compare_four}
\begin{tabular}{l|lll}
\toprule
Type & Strength & Limitation & Key Challenge \\ \midrule
\multirow{2}{*}{Signal-Transformation-Based}
& Uses domain-specific knowledge & Limited flexibility & Incorporates more brain signal features \\
& High interpretability & Struggles with nonlinear data & Ensures diversity of generated signals \\\midrule
\multirow{2}{*}{Feature-Based}
& Generates diverse samples & Lack of direct signal fidelity & Maintains feature integrity \\
& Useful for imbalanced data & Overfits to the minority & Enhances model robustness \\\midrule
\multirow{2}{*}{Model-Based}
& Handles high-dimensional data & Computationally expensive & Manages model complexity \\
& Flexible and adaptable & Prone to mode collapse & Efficient training\\\midrule
\multirow{2}{*}{Translation-Based}
& Enables multimodal integration & Complex to train and deploy & Ensures accurate cross-modal alignment \\
& Integrates more information & Hard to align modalities & Maintains cross-modal consistency\\ \bottomrule
\end{tabular}
\end{table*}

\section{Benchmark Experiments} \label{sect:benchmark}

This section details the datasets, experiments, and analyses. We benchmarked and fairly evaluated the popular brain signal generation approaches on four mainstream BCI paradigms across 11 public datasets. Code for all compared approaches is available on GitHub\footnote{https://github.com/wzwvv/DG4BCI}, serving as a benchmark codebase for brain signal generation.

\subsection{Dataset}

We conducted experiments on four BCI decoding tasks: MI, SSVEP, ESD, and AAD. Characteristics of 11 publicly available datasets are summarized in Table \ref{tab:dataset_info}.

\begin{table*}[htpb]  \centering \setlength{\tabcolsep}{1mm}
\renewcommand{\arraystretch}{1}
\caption{Summary of the five MI datasets, two seizure datasets, two SSVEP datasets, and two AAD datasets. The sampling rate and trial length refer to the preprocessed signals used in our experiments.}
\small
\label{tab:dataset_info}
\begin{tabular}{c|c|c|c|c|c|c|c}
\toprule
\multirow{2}{*}{Paradigm} & \multirow{2}{*}{Dataset} & Number of & Number of & Sampling & Trial Length & Number of & \multirow{2}{*}{Task Types} \\
& & Subjects & EEG Channels & Rate (Hz) & (seconds) & Trials & \\
\midrule
\multirow{5}{*}{MI}
& IV-2a & 9 & 22 & 250 & 4 & 1,296 & left/right hand \\
& Zhou2016 & 4 & 14 & 250 & 5 & 409 & left/right hand\\
& Blankertz2007 & 7 & 59 & 250 & 3 & 1,400 & left/right hand or left hand/right foot \\
& BNCI2014002 & 14 & 15 & 512 & 5 & 1,400 & right hand/both feet \\
& BNCI2015001 & 12 & 13 & 512 & 5 & 2,400 & right hand/both feet \\
\midrule
\multirow{2}{*}{ESD}
& CHSZ & 27 & 19 & 500 & 4 & 21,237 & normal/seizure \\
& NICU & 39 & 19 & 256 & 4 & 52,534 & normal/seizure \\
\midrule
\multirow{2}{*}{SSVEP}
& Nakanishi & 10 & 8 & 256 & 1 & 1,800 & 12 different stimuli \\
& Benchmark & 35 & 9 & 250 & 1 & 8,400 & 40 different stimuli \\
\midrule
\multirow{2}{*}{AAD}
& KUL & 16 & 64 & 128 & 2 & 2,872 & left/right attention track \\
& DTU & 18 & 64 & 64 & 2 & 2,940 & left/right attention track \\
\bottomrule
\end{tabular}
\end{table*}

\subsubsection{MI Datasets}

We adopted four publicly available MI datasets from the MOABB benchmark \cite{Jayaram2018} and BCI competitions.
\begin{itemize}
\item IV-2a \cite{tangermann2014001}: Includes EEG recordings from nine participants performing left/right hand MI tasks. Signals were collected using 22 channels at a sampling rate of 250 Hz. Only the first recording session was considered in our experiments.
\item Zhou2016 \cite{Zhou2016}: Contains MI trials from four subjects performing left/right hand imagery, recorded with 14 EEG channels at 250~Hz. Experiments were conducted using only the first session.
\item Blankertz2007 \cite{Blankertz2007MI1}: Derived from BCI Competition IV Dataset 1, comprising EEG recordings from seven subjects collected with 59 channels at 250 Hz. Two subjects performed left hand/right foot MI, while the remaining participants carried out left/right hand imagery tasks.
\item BNCI2014002 \cite{steyrl2014002}: Contains EEG trials from 14 subjects, each completing eight runs of right hand and foot MI. Signals were recorded using 15 channels at 512 Hz. The first five training runs were used.
\item BNCI2015001 \cite{Faller2012BNCI2015001}: Contains EEG trials from 12 subjects, each completing two sessions of right hand and foot MI. Signals were recorded using 13 channels at 512 Hz. The first session was used.
\end{itemize}

For the IV-2a, Zhou2016, BNCI2014002, and BNCI2015001 datasets, the standard preprocessing pipeline provided by MOABB was applied, including notch filtering and band-pass filtering. For the Blankertz2007 dataset, EEG signals were first band-pass filtered within 8-30 Hz. Trials in the interval of [0.5, 3.5] s after cue onset were extracted and downsampled to 250 Hz. Euclidean alignment (EA) \cite{He2020EA}, an effective unsupervised EEG alignment approach was applied. In the cross-subject setting, the EA reference matrix of the target subject was updated online as new test trials became available, as in \cite{Li2024T-TIME}.

\subsubsection{SSVEP Datasets}

Two publicly collected SSVEP datasets are evaluated.
\begin{itemize}
\item Nakanishi \cite{Nakanishi2015}: Comprises EEG recordings from 10 subjects under a 12-target frequency-coded visual stimulation paradigm, with stimulation frequencies ranging from 9.25 Hz to 14.75 Hz at 0.5 Hz intervals. EEG signals were acquired from 8 occipital electrodes using a BioSemi ActiveTwo system. Each subject completed 15 blocks of trials, with a 4-second trial length.
\item Benchmark \cite{benchmark2017}: Consists of 64-channel EEG recordings from 35 healthy subjects. The experiment employed a cue-guided virtual keyboard with 40 visual targets encoded using a joint frequency-phase modulation scheme. Stimulation frequencies ranged from 8.0 Hz to 15.8 Hz with a 0.2 Hz step and a fixed phase difference of $0.5\pi$. Each subject completed six blocks of trials, with a 5-second trial length.
\end{itemize}

For Nakanishi, the data were downsampled to 256 Hz, and then split into 1-second length trials for analysis. For Benchmark, EEG signals were downsampled to 250 Hz, and then split into 1-second length trials for analysis. Nine posterior electrodes (Pz, PO3, PO5, PO4, PO6, POz, O1, Oz, and O2) were selected following \cite{ravi2020CCNN}.

\subsubsection{ESD Datasets}

Two clinical seizure detection datasets are evaluated.
\begin{itemize}
\item CHSZ \cite{Wang2023TASA}: Comprises EEG recordings from 27 pediatric patients (aged 3 months to 10 years), acquired using 19 unipolar channels at sampling rates of 500 or 1000 Hz. The 1,000 Hz trials were downsampled to 500 Hz for consistency. All seizure events were annotated by certified neurologists, with precise labeling of seizure onset and offset.
\item NICU \cite{Stevenson2019}: A large-scale neonatal EEG corpus collected in the neonatal intensive care unit of Helsinki University Hospital. Annotations were provided independently by three clinical experts with a temporal resolution of 1s. For our experiments, we adopt a consensus subset consisting of 39 neonates with confirmed seizure activity.
\end{itemize}

For preprocessing, EEG signals in both seizure datasets were recorded with 19 unipolar electrodes placed according to the international 10-20 system, and 18 bipolar channels were derived from them \cite{Stevenson2019}. Each bipolar channel was preprocessed using a 50 Hz notch filter and a 0.5-50 Hz band-pass filter, after which the continuous signals were segmented into 4-second trials \cite{Wang2023TASA}.

\subsubsection{AAD Datasets}

We further evaluate on two widely used public AAD datasets, namely the KUL and DTU datasets. Both datasets record EEG responses elicited in multi-speaker listening scenarios, where subjects are required to selectively attend to one target speaker.
\begin{itemize}
\item KUL \cite{das2016effect}: Consists of 64-channel EEG recordings from 16 normal-hearing participants, acquired at a sampling rate of 8,192 Hz. During the collection, subjects listened to mixtures of two concurrent speech streams and were instructed to attend to one of them. Two spatial presentation conditions were included, a dichotic condition with one speaker per ear and a spatialized condition using head-related transfer functions to simulate sound sources located at $\pm90^\circ$. Each participant completed eight trials, each lasting approximately six minutes. The EEG trials were further high-pass filtered at 0.5 Hz and downsampled to 128 Hz.
\item DTU \cite{fuglsang2017noise}: 64-channel EEG data were collected from 18 normal-hearing subjects at a sampling rate of 512 Hz. The experimental paradigm similarly involved selective attention to one of two simultaneous speakers presented at $\pm60^\circ$ relative to the listener. Speech stimuli were drawn from Danish audiobooks narrated by both male and female speakers and delivered through ER-2 earphones at 60~dB. Each subject participated in 60 trials with a 50-second trial length.
\end{itemize}

For KUL, EEG trials were high-pass filtered at 0.5 Hz and downsampled to 128 Hz. Artifacts were removed using the MWF filtering approach \cite{somers2018generic}. For DTU, The signals were high-pass filtered at 0.1 Hz, notch filtered at 50 Hz, and then downsampled to 64 Hz. For both datasets, signals were segmented into 2-second trials.

\subsection{Generation Algorithms}

We benchmarked 8 representative signal-transformation-based EEG data generation approaches introduced in Table~\ref{tab:kbDG}, including Flip, Noise, Scale, Fshift, FSurr, HS, CR, and DWTaug.
\begin{itemize}
\item None: No data generation is applied.
\item Flip \cite{Zhang2022MSDT}: Trial amplitudes are inverted in the time domain.
\item Noise \cite{Wang2018}: Uniform noise is added to EEG trials in the time domain.
\item Scale \cite{Zhang2022MSDT}: Trial amplitudes are scaled by a coefficient close to one in the time domain.
\item FShift \cite{Zhang2022MSDT}: Trial frequencies are shifted using the Hilbert transform \cite{Freeman2007}.
\item FSurr \cite{Schwabedal2018}: Fourier phase components are replaced with random values.
\item HS \cite{pei2021hs}: Trials are split by hemispheric channels and randomly recombined within the same class.
\item CR \cite{Wang2024CR}: Symmetric left-right hemisphere channels are exchanged, with labels swapped for left/right hand MI tasks.
\item DWTaug \cite{Wang2025CSDA}: Trials are decomposed using DWT, then reassembled coefficients and reconstructed to generate new trials. In the implementation, we performed DWTaug on the raw signals and their reversed signals, augmenting the raw data to three times. The new implementation is introduced in the repository\footnote{https://github.com/wzwvv/CSDA/blob/main/DWTaug-reverse.py}.
\end{itemize}

Besides, 9 representative generation models were benchmarked, selected from model families that can be implemented and compared within a unified experimental setting. Autoregressive and diffusion models were excluded due to their substantially different training objectives, sampling procedures, and computational requirements, which complicate direct comparison under a single protocol.
\begin{itemize}
\item Vanilla CNN: The generator comprises a shallow CNN architecture coupled with its inverse counterpart for EEG synthesis, and is trained with a combination of classification and $\ell_1$ losses.
\item Vanilla CNN-Trans.: Building upon the vanilla CNN baseline, the generator integrates CNN blocks with a Transformer encoder \cite{wang2025dbconformer}.
\item Vanilla CNN-LSTM: Extending the vanilla CNN architecture, the generator incorporates an LSTM layer alongside CNN modules, as in \cite{zeng2024task}.
\item GAN-based CNN: In this adversarial variant, the backbone includes a discriminator and is optimized using a classification loss, an $\ell_1$ loss, and the standard GAN discriminative loss \cite{goodfellow2014generative}.
\item GAN-based CNN-Trans.: The generator combines CNN blocks with a Transformer encoder \cite{wang2025dbconformer}, and is trained with classification, $\ell_1$, and discriminative losses \cite{goodfellow2014generative}.
\item GAN-based CNN-LSTM: The generator merges CNN modules with an LSTM layer \cite{zeng2024task}, optimized via classification, $\ell_1$, and adversarial losses \cite{goodfellow2014generative}.
\item VAE-based CNN: In addition to the classification and $\ell_1$ losses, a latent alignment loss between the latent representations of real and generated data is introduced to guide training, following \cite{tibermacine2025conditional}.
\item VAE-based CNN-Trans.: The generator integrates CNN blocks with a Transformer encoder \cite{wang2025dbconformer} and is optimized using classification, $\ell_1$, and latent alignment losses \cite{tibermacine2025conditional}.
\item VAE-based CNN-LSTM: The generator combines CNN blocks with an LSTM layer \cite{zeng2024task}, trained with classification, $\ell_1$, and latent alignment losses \cite{tibermacine2025conditional}.
\end{itemize}

\subsection{Decoding Algorithms}

Different types of EEG decoding algorithms, including traditional machine learning approaches and deep neural networks, were benchmarked in the experiments:
\begin{itemize}
\item EEGNet \cite{Lawhern2018EEGNet} is a compact CNN tailored for EEG classification. It includes two key convolutional blocks: a temporal convolution for capturing frequency-specific features, followed by a depthwise spatial convolution. A separable convolution and subsequent pointwise convolution are designed to enhance spatio-temporal representations.
\item SCNN \cite{deepshallow2017} is a shallow CNN inspired by the filter bank common spatial pattern, utilizing temporal and spatial convolutions in two stages to efficiently extract discriminative EEG features.
\item DCNN \cite{deepshallow2017} is a deeper version of SCNN with larger parameters, consisting of four convolutional blocks with max pooling layers.
\item IFNet \cite{wang2023ifnet} extends the spectral-spatial approach by decomposing EEG into multiple frequency bands (e.g., 4-16 Hz, 16-40 Hz). For each band, it applies 1D spatial and temporal convolutions, followed by feature concatenation and a fully connected layer for classification.
\item EEGWaveNet \cite{EEGWaveNet2022} introduces a multi-scale temporal CNN framework that employs channel-wise depthwise filters to learn representations at multiple temporal resolutions from each EEG channel independently.
\item Filter bank canonical correlation analysis (FBCCA) \cite{chen2015fbcca} is a widely used SSVEP decoding approach that extends standard canonical correlation analysis \cite{lin2006cca} by incorporating a filter bank. For both datasets, a filter bank composed of 3 sub-bands was used.
\item CCNN \cite{ravi2020CCNN} is designed with convolutional layers for spatial and spectral feature extraction, followed by fully connected layers for classification. By explicitly modeling both magnitude and phase information, CCNN provides an effective baseline for SSVEP recognition.
\item SSVEPNet \cite{Pan2022} combines one-dimensional CNN layers with an LSTM module for SSVEP detection. CNN layers extract local temporal features, and the LSTM layer captures long-range temporal dependencies.
\item DARNet \cite{yan2024darnet} is a dual attention refinement network for the AAD task, consisting of the spatiotemporal construction module, dual attention refinement module, feature fusion, and classifier module.
\item DBPNet \cite{ni2024dbpnet} a dual-branch parallel network with temporal-frequency fusion for AAD, which consists of the temporal attentive branch and the frequency residual branch.
\item DBConformer \cite{wang2025dbconformer} is a dual-branch convolutional Transformer model for EEG decoding, comprising a temporal branch T-Conformer and a spatial branch S-Conformer to simultaneously learn the temporal dynamics and spatial patterns of EEG data.
\end{itemize}

\subsection{Evaluation Settings}

\subsubsection{Evaluation Scenarios} 

To evaluate the performance of generation approaches under different scenarios, we conducted experiments under the within-subject and cross-subject settings, as in \cite{wang2025dbconformer,wang2025cst}. The cross-subject scenario evaluates the generalization performance of each approach, while the within-subject scenario mimics real-time deployment. Specifically, the MI and ESD experiments were conducted under a cross-subject setting, where EEG trials from a single subject were held out as the test set and trials from all remaining subjects were combined for training. SSVEP and AAD experiments were under the within-subject setting. A 5-fold or 6-fold cross-validation was performed on SSVEP, with the dataset divided chronologically into 5 or 6 equal parts, each with balanced classes. For the Nakanishi dataset, one fold was used for training and four for testing in each iteration when applying signal-transformation-based generation approaches. In contrast, four folds were used for training and one for testing when applying the model-based generation, as generative models require more training data. For the Benchmark dataset, five folds were used for training and one fold for testing, since the trial number per class is 6. For AAD, trials from each subject were split according to recording time. The first 90\% of trials were used for training, and the remaining 10\% for testing.

\subsubsection{Evaluation Metrics}

Accuracy was used to evaluate the performance on MI, SSVEP, and AAD paradigms. To evaluate the classification performance on class-imbalanced seizure datasets, the balanced classification accuracy (BCA), defined as the average of recall obtained on each class, was calculated in the ESD paradigm.

\subsection{Implementation Details}

\subsubsection{Training Settings} For all paradigms, experiments were repeated five times with a seed list \{1, 2, 3, 4, 5\}, and the average results were reported. For MI classification, models were trained for 100 epochs using the Adam optimizer with learning rate $1\times10^{-3}$. Batch sizes were set to 32 for baselines and 64 for data generation approaches. For ESD, the training epoch 100, learning rate $10^{-3}$, and batch size 256 were used for both datasets and both backbone nets. For SSVEP detection, the batch size was set to 36, the training epoch was 300, and the learning rate was $10^{-3}$. Weight decay with a coefficient of $10^{-4}$ was applied to mitigate overfitting. For AAD, batch size $32$, training epoch $200$ using early stopping with patience 10, learning rate $5\times10^{-4}$, and weight decay with a coefficient of $3\times10^{-4}$ were applied on all backbones.
\subsubsection{Generative Models Settings} For model-based generation, the experiments consisted of two stages. The generative models (nine variants in total) were first pretrained using the training data to learn the underlying distribution of brain signals. Then, the generated signals were concatenated with the original training signals to train downstream classifiers, whose performance was evaluated on the held-out test data. Different objective functions were used for optimization. For the vanilla models, an $\ell_1$ reconstruction loss was used to constrain the generated signals. For GAN-based models, the adversarial loss was introduced, where a discriminator was trained to distinguish real and generated samples, and a domain classifier was employed. For VAE-based models, the standard variational objective was adopted with the KL divergence constraining the latent distribution to match a standard normal distribution. The trade-off parameter between the $\ell_1$ loss and alignment loss was 1 for both GAN-based and VAE-based models.

\subsubsection{Parameter Settings} The noise rate was set to 2 for Noise, the scaling rate was set to 0.05 for Scale, and the frequency scaling rate was 0.2 for FShift, following \cite{Freer2020}. For FSurr, the probability was 1, and the phase noise magnitude was 0.5, following \cite{Schwabedal2018}. For DWTaug, the number of decomposition levels was set to 2, as in \cite{Wang2025CSDA}. Note that Flip, HS, and CR do not require any hyperparameters.

\subsection{Main Results}

\subsubsection{Results on MI}

To evaluate the effectiveness of signal-transformation-based generation approaches, we compared eight representative strategies under a cross-subject setting on five MI datasets. Five widely used decoding models were adopted as backbones, including SCNN, DCNN, EEGNet, IFNet, and the CNN-Transformer hybrid model DBConformer, following \cite{wang2025mvcnet}. The results are summarized in Table~\ref{tab:mi_results}. Observe that:
\begin{itemize}
\item Overall, most signal-transformation-based generation strategies improved the None baseline, but their gains were clearly different. DWTaug achieved the best average performance for four out of five backbones and yielded the highest overall accuracy. For example, it improved SCNN from 72.41\% to 76.30\% and DCNN from 73.54\% to 77.21\%. This suggested that MI decoding benefits from transformations that preserve task-related signal structures while increasing data diversity. In contrast, some perturbations were less reliable: FSurr substantially degraded SCNN and DCNN, and HS caused a large drop with DCNN. These failure cases indicated that disrupting Fourier phase or spatial organization may damage discriminative MI patterns, rather than simply increasing the variability.
\item The effectiveness of augmentation also depended on the decoding backbone. DBConformer achieved the highest overall performance, reaching 79.50\% with DWTaug, while IFNet also showed relatively stable gains across several transformations. Compared with shallow CNN backbones, stronger temporal modeling architectures are less sensitive to harmful perturbations and can better exploit informative synthetic samples. This suggested that the observed improvements may not solely reflect the realism of generated signals, but also the interaction between the induced invariances and the feature extraction capacity of downstream decoders.
\end{itemize}

\begin{table*}[htpb]
\centering
\small
\setlength{\tabcolsep}{0.8mm}
\renewcommand\arraystretch{1}
\caption{Average classification accuracies (\%) on five MI datasets with five decoding models under the cross-dataset setting. The best average performance of each network is marked in bold, and the second best by an underline.}  \label{tab:mi_results}
\begin{tabular}{c|c|ccccccccc}   \toprule
Backbone & Dataset & None & Flip & Noise & Scale & FShift & FSurr & HS & CR & DWTaug \\
\midrule
\multirow{6.5}{*}{SCNN}
& IV-2a & 72.22$_{\pm1.03}$ & \underline{75.10}$_{\pm0.95}$ & 74.44$_{\pm0.74}$ & 74.39$_{\pm0.51}$ & 74.95$_{\pm0.95}$ & 65.92$_{\pm1.24}$ & 73.37$_{\pm1.05}$ & 74.76$_{\pm1.10}$ & \textbf{75.56}$_{\pm0.93}$ \\
& Zhou2016 & 81.97$_{\pm1.63}$ & 81.82$_{\pm0.74}$ & 82.62$_{\pm1.20}$ & \underline{83.67}$_{\pm0.91}$ & 81.67$_{\pm0.61}$ & 63.04$_{\pm1.20}$ & 81.57$_{\pm0.58}$ & 81.87$_{\pm0.78}$ & \textbf{84.92}$_{\pm1.32}$ \\
& Blankertz2007 & 70.04$_{\pm0.79}$ & \underline{74.10}$_{\pm0.90}$ & 73.59$_{\pm0.90}$ & 71.96$_{\pm0.87}$ & 73.57$_{\pm0.44}$ & 61.51$_{\pm0.88}$ & 73.38$_{\pm0.63}$ & 73.89$_{\pm0.66}$ & \textbf{74.69}$_{\pm1.17}$ \\
& BNCI2014002 & 68.13$_{\pm0.74}$ & 72.10$_{\pm0.37}$ & \textbf{72.97}$_{\pm0.41}$ & 72.36$_{\pm0.74}$ & 72.28$_{\pm0.70}$ & 66.74$_{\pm1.14}$ & 71.72$_{\pm0.67}$ & \underline{72.43}$_{\pm0.45}$ & 72.21$_{\pm1.01}$ \\
& BNCI2015001 & 69.71$_{\pm0.67}$ & \underline{69.80}$_{\pm0.89}$ & 69.31$_{\pm1.00}$ & 68.98$_{\pm0.82}$ & 69.35$_{\pm0.90}$ & 68.60$_{\pm0.65}$ & 68.32$_{\pm0.96}$ & 69.50$_{\pm0.73}$ & \textbf{74.11}$_{\pm0.29}$ \\
\cmidrule{2-11}
& Average & 72.41 & 74.58 & \underline{74.59} & 74.27 & 74.36 & 65.16 & 73.67 & 74.49 & \textbf{76.30} \\
\midrule
\multirow{6.5}{*}{DCNN}
& IV-2a & 73.21$_{\pm1.79}$ & 76.13$_{\pm0.87}$ & 72.66$_{\pm0.82}$ & 76.13$_{\pm1.00}$ & \underline{76.42}$_{\pm1.24}$ & 62.29$_{\pm0.66}$ & 50.05$_{\pm1.48}$ & \textbf{77.80}$_{\pm0.45}$ & 73.92$_{\pm0.81}$ \\
& Zhou2016 & 82.91$_{\pm0.85}$ & 84.01$_{\pm1.21}$ & 83.27$_{\pm0.44}$ & 83.22$_{\pm1.30}$ & 83.69$_{\pm1.13}$ & 78.76$_{\pm1.73}$ & 53.88$_{\pm1.43}$ & \underline{84.22}$_{\pm1.26}$ & \textbf{85.55}$_{\pm1.10}$ \\
& Blankertz2007 & 71.44$_{\pm0.78}$ & 71.31$_{\pm0.76}$ & 71.06$_{\pm1.32}$ & 71.80$_{\pm0.46}$ & 70.87$_{\pm0.52}$ & 67.39$_{\pm0.76}$ & 50.72$_{\pm0.42}$ & \underline{72.10}$_{\pm1.50}$ & \textbf{72.32}$_{\pm0.62}$ \\
& BNCI2014002 & 69.74$_{\pm0.94}$ & 74.76$_{\pm0.97}$ & 74.74$_{\pm1.12}$ & 74.98$_{\pm0.80}$ & \underline{75.14}$_{\pm0.87}$ & 72.23$_{\pm0.46}$ & 54.41$_{\pm2.60}$ & 74.35$_{\pm1.20}$ & \textbf{77.54}$_{\pm0.75}$ \\
& BNCI2015001 & 70.42$_{\pm0.68}$ & 73.65$_{\pm0.82}$ & 74.20$_{\pm0.41}$ & \underline{74.32}$_{\pm0.86}$ & 74.12$_{\pm0.75}$ & 73.86$_{\pm0.62}$ & 61.78$_{\pm2.59}$ & 73.93$_{\pm0.35}$ & \textbf{76.71}$_{\pm0.45}$ \\
\cmidrule{2-11}
& Average & 73.54 & 75.97 & 75.19 & 76.09 & 76.05 & 70.91 & 54.17 & \underline{76.48} & \textbf{77.21} \\
\midrule
\multirow{6.5}{*}{EEGNet}
& IV-2a & 73.64$_{\pm1.14}$ & 73.72$_{\pm1.14}$ & 73.98$_{\pm1.06}$ & 73.07$_{\pm0.84}$ & 73.21$_{\pm0.76}$ & 74.38$_{\pm0.89}$ & 71.58$_{\pm0.88}$ & \textbf{75.77}$_{\pm1.58}$ & \underline{75.35}$_{\pm1.39}$ \\
& Zhou2016 & 83.22$_{\pm1.73}$ & 81.19$_{\pm2.39}$ & 84.16$_{\pm0.96}$ & 83.99$_{\pm0.83}$ & 82.94$_{\pm1.99}$ & 83.82$_{\pm1.18}$ & 80.18$_{\pm2.52}$ & \underline{84.82}$_{\pm1.36}$ & \textbf{85.00}$_{\pm1.17}$ \\
& Blankertz2007 & 71.17$_{\pm0.87}$ & 69.86$_{\pm1.49}$ & 71.87$_{\pm0.55}$ & 71.70$_{\pm0.73}$ & 70.96$_{\pm0.82}$ & 69.82$_{\pm0.56}$ & 68.31$_{\pm2.68}$ & \textbf{74.97}$_{\pm0.96}$ & \underline{72.17}$_{\pm1.37}$ \\
& BNCI2014002 & 72.86$_{\pm0.38}$ & \underline{73.75}$_{\pm1.31}$ & 72.49$_{\pm0.69}$ & 72.59$_{\pm0.68}$ & 73.51$_{\pm1.07}$ & 72.21$_{\pm0.97}$ & 69.99$_{\pm2.37}$ & 72.43$_{\pm0.70}$ & \textbf{74.13}$_{\pm0.83}$ \\
& BNCI2015001 & 71.89$_{\pm0.70}$ & 71.96$_{\pm1.11}$ & 72.28$_{\pm1.02}$ & 71.73$_{\pm1.30}$ & 73.11$_{\pm1.31}$ & \underline{73.21}$_{\pm1.20}$ & 70.91$_{\pm2.17}$ & 72.21$_{\pm0.93}$ & \textbf{76.92}$_{\pm0.59}$ \\
\cmidrule{2-11}
& Average & 74.56 & 74.10 & 74.96 & 74.62 & 74.75 & 74.69 & 72.19 & \underline{76.04} & \textbf{76.71} \\
\midrule
\multirow{6.5}{*}{IFNet}
& IV-2a & 74.52$_{\pm0.75}$ & 74.81$_{\pm1.02}$ & \textbf{75.82}$_{\pm0.41}$ & 75.45$_{\pm0.46}$ & \underline{75.47}$_{\pm0.63}$ & 74.98$_{\pm0.72}$ & 73.56$_{\pm1.10}$ & 74.29$_{\pm0.59}$ & 74.65$_{\pm0.54}$ \\
& Zhou2016 & \underline{86.21}$_{\pm0.99}$ & 85.71$_{\pm0.41}$ & 85.66$_{\pm0.57}$ & 86.17$_{\pm0.49}$ & 85.91$_{\pm0.69}$ & 79.46$_{\pm1.33}$ & 85.38$_{\pm0.95}$ & 86.05$_{\pm0.62}$ & \textbf{87.03}$_{\pm0.46}$ \\
& Blankertz2007 & 73.43$_{\pm1.06}$ & 76.16$_{\pm0.43}$ & 76.16$_{\pm0.37}$ & 75.83$_{\pm0.52}$ & 76.00$_{\pm0.74}$ & 75.09$_{\pm0.86}$ & 72.99$_{\pm1.25}$ & \textbf{79.36}$_{\pm1.33}$ & \underline{76.26}$_{\pm0.92}$ \\
& BNCI2014002 & 73.90$_{\pm0.65}$ & 75.33$_{\pm1.02}$ & 75.93$_{\pm0.69}$ & \underline{75.96}$_{\pm0.57}$ & 75.83$_{\pm1.13}$ & 75.14$_{\pm0.71}$ & 75.89$_{\pm0.64}$ & 71.44$_{\pm0.45}$ & \textbf{76.00}$_{\pm0.79}$ \\
& BNCI2015001 & 72.73$_{\pm1.17}$ & 72.96$_{\pm0.60}$ & 72.43$_{\pm0.85}$ & 72.54$_{\pm0.73}$ & \underline{73.13}$_{\pm0.53}$ & 72.30$_{\pm0.43}$ & 71.32$_{\pm1.87}$ & 70.83$_{\pm0.76}$ & \textbf{78.16}$_{\pm0.52}$ \\
\cmidrule{2-11}
& Average & 76.16 & 76.99 & 77.20 & 77.19 & \underline{77.27} & 75.39 & 75.83 & 76.39 & \textbf{78.42} \\
\midrule
\multirow{6.5}{*}{DBConformer}
& IV-2a & \textbf{77.67}$_{\pm0.61}$ & 76.22$_{\pm1.19}$ & 76.54$_{\pm0.82}$ & 76.27$_{\pm0.43}$ & 76.03$_{\pm1.06}$ & 74.52$_{\pm1.48}$ & 73.01$_{\pm0.83}$ & 76.06$_{\pm0.46}$ & \underline{77.29}$_{\pm1.35}$ \\
& Zhou2016 & 85.37$_{\pm0.94}$ & 85.54$_{\pm1.10}$ & 85.58$_{\pm0.59}$ & \underline{86.53}$_{\pm0.73}$ & 85.82$_{\pm0.92}$ & 81.92$_{\pm1.65}$ & 86.07$_{\pm0.75}$ & 86.19$_{\pm0.85}$ & \textbf{86.73}$_{\pm0.77}$ \\
& Blankertz2007 & 76.33$_{\pm0.71}$ & 76.44$_{\pm1.26}$ & 76.5$_{\pm1.11}$ & 76.09$_{\pm1.01}$ & 76.29$_{\pm1.10}$ & \underline{77.44}$_{\pm1.12}$ & 75.73$_{\pm0.85}$ & \textbf{78.99}$_{\pm0.65}$ & 76.84$_{\pm0.93}$ \\
& BNCI2014002 & 77.17$_{\pm0.78}$ & 76.96$_{\pm0.12}$ & 77.01$_{\pm0.86}$ & \underline{77.90}$_{\pm0.82}$ & 77.24$_{\pm0.44}$ & 77.64$_{\pm0.94}$ & 73.04$_{\pm0.96}$ & 73.97$_{\pm0.75}$ & \textbf{78.46}$_{\pm0.56}$ \\
& BNCI2015001 & 75.90$_{\pm0.90}$ & 75.45$_{\pm0.54}$ & \underline{76.00}$_{\pm0.89}$ & 75.45$_{\pm0.30}$ & 75.17$_{\pm0.60}$ & 73.46$_{\pm1.13}$ & 73.38$_{\pm1.07}$ & 75.21$_{\pm0.82}$ & \textbf{78.17}$_{\pm0.71}$ \\
\cmidrule{2-11}
& Average & \underline{78.49} & 78.12 & 78.33 & 78.45 & 78.11 & 77.00 & 76.25 & 78.08 & \textbf{79.50} \\
\bottomrule
\end{tabular}
\end{table*}

\subsubsection{Results on ESD}

Classification results on two seizure datasets are shown in Table~\ref{tab:esd_results}. Observe that:
\begin{itemize}
\item Compared with MI, signal-transformation-based generation showed more selective benefits for ESD. Several transformations degraded the None baseline, especially HS and Flip, suggesting that seizure-related brain signal patterns are sensitive to temporal and spatial distortions. This is reasonable because seizure detection relies on preserving pathological waveform morphology and rhythmic discharge patterns; inappropriate transformations may introduce unrealistic variations rather than useful diversity.
\item CR achieved the best average performance on both CHSZ and NICU, improving BCA from 79.99\% to 81.12\% and from 62.04\% to 64.04\%, respectively. Unlike directly altering signal amplitude or temporal structure, CR mainly modified the spatial configuration by swapping symmetric hemispheric channels. This indicated that spatially plausible transformations can enrich inter-hemispheric variability while preserving seizure-related temporal morphology, highlighting the importance of spatial information in ESD.
\end{itemize}

\begin{table*}[htpb]
\centering
\small
\setlength{\tabcolsep}{3mm}
\renewcommand\arraystretch{1}
\caption{Average classification BCAs (\%) on two seizure datasets with EEGNet and EEGWaveNet models under the cross-subject setting. The best average performance of each network is marked in bold, and the second best by an underline.}  \label{tab:esd_results}
\begin{tabular}{c|c|ccccccc}   \toprule
Dataset & Backbone & None & Flip & Noise & Scale & FShift & HS & CR \\
\midrule
\multirow{3.5}{*}{CHSZ}
& EEGNet & \underline{81.92}$_{\pm0.76}$ & 79.46$_{\pm0.78}$ & 80.63$_{\pm0.81}$ & 81.74$_{\pm0.31}$ & 80.43$_{\pm0.93}$ & 80.40$_{\pm1.19}$ & \textbf{82.77}$_{\pm0.40}$ \\
& EEGWaveNet & 78.05$_{\pm0.34}$ & \underline{79.03}$_{\pm0.70}$ & 78.21$_{\pm0.97}$ & 79.02$_{\pm0.02}$ & 77.65$_{\pm1.31}$ & 75.25$_{\pm1.09}$ & \textbf{79.47}$_{\pm0.57}$ \\
\cmidrule{2-9}
& Average & 79.99 & 79.25 & 79.42 & \underline{80.38} & 79.04 & 77.83 & \textbf{81.12} \\
\midrule
\multirow{3.5}{*}{NICU}
& EEGNet & \underline{61.22}$_{\pm0.54}$ & 60.43$_{\pm0.24}$ & 58.88$_{\pm0.29}$ & 58.69$_{\pm0.43}$ & 58.79$_{\pm0.51}$ & 57.17$_{\pm0.46}$ & \textbf{62.03}$_{\pm0.40}$ \\
& EEGWaveNet & 62.86$_{\pm0.68}$ & 61.17$_{\pm0.47}$ & 62.88$_{\pm0.50}$ & \underline{65.12}$_{\pm0.71}$ & 63.38$_{\pm0.53}$ & 63.39$_{\pm0.23}$ & \textbf{66.04}$_{\pm0.16}$ \\
\cmidrule{2-9}
& Average & \underline{62.04} & 60.80 & 60.88 & 61.91 & 61.09 & 60.28 & \textbf{64.04} \\
\bottomrule
\end{tabular}
\end{table*}

\subsubsection{Results on SSVEP}

Classification results on two SSVEP datasets are presented in Table~\ref{tab:ssvep_results}. The traditional FBCCA and three deep models were used for decoding. Observe that:
\begin{itemize}
\item Overall, SSVEP benefited most from transformations that preserve frequency-locked periodic responses. DWTaug achieved the best average performance on both datasets, improving accuracy from 68.58\% to 71.76\% on Nakanishi and from 71.07\% to 74.69\% on Benchmark. This suggested that wavelet decomposition and recombination can enrich discriminative frequency components while maintaining the intrinsic periodic structure required for SSVEP decoding.
\item Different transformations showed clear paradigm-specific effects. Noise and Scale provided moderate improvements in several cases, suggesting that mild perturbations can act as regularization without changing the target frequency structure. In contrast, FShift often degraded performance, especially for FBCCA, because shifting frequency components may weaken the correspondence between EEG responses and stimulation frequencies. Flip caused the most severe performance drop across all models. Mathematically, Flip multiplies the EEG signal by $-1$, inducing a global $\pi$ phase shift while preserving the amplitude spectrum. Since SSVEP decoding relies heavily on phase synchronization between EEG responses and stimulus frequencies, phase inversion disrupts this key discriminative structure, leading to substantial degradation.
\end{itemize}

\begin{table*}[htpb]
\centering
\small
\setlength{\tabcolsep}{3.7mm}
\renewcommand\arraystretch{1}
\caption{Average classification accuracies (\%) on two SSVEP datasets with four decoding backbones under the within-subject setting. The best average performance of each network is marked in bold, and the second best by an underline.}  \label{tab:ssvep_results}
\begin{tabular}{c|c|cccccc}   \toprule
Dataset & Backbone & None & Flip & Noise & Scale & FShift & DWTaug \\
\midrule
\multirow{5.5}{*}{Nakanishi}
& FBCCA & \underline{68.06}$_{\pm0.68}$ & 7.90$_{\pm0.57}$ & 67.99$_{\pm0.69}$ & \underline{68.06}$_{\pm0.68}$ & 60.68$_{\pm2.21}$ & \textbf{72.86}$_{\pm3.37}$ \\
& EEGNet & 45.69$_{\pm3.71}$ & 35.90$_{\pm1.13}$ & \underline{49.88}$_{\pm2.79}$ & 49.14$_{\pm2.82}$ & 48.63$_{\pm3.94}$ & \textbf{51.93}$_{\pm2.70}$ \\
& CCNN & 70.54$_{\pm1.74}$ & 69.14$_{\pm3.49}$ & \underline{70.65}$_{\pm3.33}$ & 70.81$_{\pm3.41}$ & 70.08$_{\pm2.27}$ & \textbf{71.36}$_{\pm3.10}$ \\
& SSVEPNet & 90.03$_{\pm1.20}$ & 82.89$_{\pm0.87}$ & 90.18$_{\pm1.49}$ & \underline{90.22}$_{\pm1.39}$ & 88.65$_{\pm1.38}$ & \textbf{90.88}$_{\pm1.08}$ \\
\cmidrule{2-8}
& Average & 68.58$_{\pm1.83}$ & 48.96$_{\pm1.52}$ & \underline{69.68}$_{\pm2.08}$ & 69.56$_{\pm2.08}$ & 67.01$_{\pm2.45}$ & \textbf{71.76}$_{\pm2.56}$ \\
\midrule
\multirow{5.5}{*}{Benchmark}
& FBCCA & \underline{63.08}$_{\pm1.06}$ & 2.38$_{\pm0.42}$ & 62.83$_{\pm1.53}$ & \underline{63.08}$_{\pm1.06}$ & 54.20$_{\pm3.07}$ & \textbf{71.42}$_{\pm0.66}$ \\
& EEGNet & 61.76$_{\pm0.75}$ & 30.52$_{\pm1.79}$ & 62.14$_{\pm1.14}$ & \textbf{62.41}$_{\pm1.04}$ & 56.35$_{\pm1.38}$ & \underline{62.17}$_{\pm0.63}$ \\
& CCNN & 71.43$_{\pm1.32}$ & 70.17$_{\pm1.28}$ & 74.26$_{\pm1.67}$ & 74.40$_{\pm1.65}$ & \underline{75.05}$_{\pm1.31}$ & \textbf{75.70}$_{\pm1.72}$ \\
& SSVEPNet & 88.04$_{\pm1.75}$ & 83.52$_{\pm1.51}$ & 87.22$_{\pm3.58}$ & \textbf{90.15}$_{\pm2.14}$ & 85.73$_{\pm2.42}$ & \underline{89.48}$_{\pm0.79}$ \\
\cmidrule{2-8}
& Average & 71.07$_{\pm1.22}$ & 46.65$_{\pm1.25}$ & 71.61$_{\pm1.98}$ & \underline{72.51}$_{\pm1.47}$ & 67.83$_{\pm2.05}$ & \textbf{74.69}$_{\pm0.95}$ \\
\bottomrule
\end{tabular}
\end{table*}

\subsubsection{Results on AAD}

Classification results on two AAD datasets are presented in Table~\ref{tab:aad_results}. Observe that:
\begin{itemize}
\item Overall, AAD showed clear benefits from frequency domain transformations. FShift achieved the best average performance on KUL and tied for the best on DTU, while DWTaug consistently ranked among the top approaches. This suggested that moderate spectral perturbations can increase variability in neural tracking patterns without severely disrupting the temporal correspondence between EEG responses and attended speech, which is essential for AAD.
\item The effects of generation were strongly backbone-dependent. For IFNet and DBConformer, frequency domain approaches such as FShift, FSurr, and DWTaug generally yielded the most reliable gains. In contrast, Flip was harmful for IFNet but achieved the highest accuracy with DARNet on both datasets. Since DARNet explicitly refines spatiotemporal dependencies through dual attention modules, polarity inversion may preserve the attention-relevant correlation while providing additional training variability. This indicated that whether a transformation is useful depends not only on the paradigm but also on the invariances it captures.
\item These results further suggest that improvements in AAD do not necessarily come from generating more realistic signals, but may arise from inducing useful neural speech tracking. Therefore, data generation strategies for AAD should be selected jointly with the decoding architecture rather than treated as universally beneficial preprocessing steps.
\end{itemize}

\begin{table*}[htpb]
\centering
\setlength{\tabcolsep}{2mm}
\small
\renewcommand\arraystretch{1}
\caption{Average classification accuracies (\%) on two AAD datasets with IFNet, DBConformer, and DARNet models under the within-subject setting. The best average performance of each network is marked in bold, and the second best by an underline.}  \label{tab:aad_results}
\begin{tabular}{c|c|cccccccc}   \toprule
Dataset & Backbone & None & Flip & Noise & Scale & FShift & FSurr & CR & DWTaug \\
\midrule
\multirow{4.5}{*}{KUL}
& IFNet & 90.46$_{\pm0.25}$ & 84.76$_{\pm2.13}$ & 90.73$_{\pm0.36}$ & 90.92$_{\pm0.27}$ & \textbf{92.77}$_{\pm0.49}$ & \underline{91.43}$_{\pm0.30}$ & 91.17$_{\pm0.46}$ & 91.39$_{\pm0.19}$ \\
& DBConformer & 84.56$_{\pm0.36}$ & 84.26$_{\pm1.12}$ & 85.20$_{\pm0.66}$ & 85.38$_{\pm0.94}$ & \textbf{88.56}$_{\pm0.34}$ & 85.05$_{\pm0.36}$ & 85.63$_{\pm0.43}$ & \underline{86.35}$_{\pm0.66}$ \\
& DARNet & 89.60$_{\pm0.72}$ & \textbf{93.58}$_{\pm0.30}$ & 89.81$_{\pm0.43}$ & 89.85$_{\pm0.30}$ & \underline{90.80}$_{\pm0.45}$ & 89.89$_{\pm0.38}$ & 89.67$_{\pm0.59}$ & 90.45$_{\pm0.45}$ \\
\cmidrule{2-10}
& Average & 88.21 & 87.53 & 88.58 & 88.72 & \textbf{90.71} & 88.79 & 88.82 & \underline{89.40} \\
\midrule
\multirow{4.5}{*}{DTU}
& IFNet & 82.25$_{\pm0.75}$ & 76.77$_{\pm0.63}$ & 81.98$_{\pm0.54}$ & 82.48$_{\pm0.48}$ & \underline{83.09}$_{\pm0.45}$ & \textbf{83.32}$_{\pm0.39}$ & 80.30$_{\pm0.32}$ & 82.50$_{\pm0.34}$ \\
& DBConformer & 80.42$_{\pm0.32}$ & 79.93$_{\pm0.81}$ & 80.74$_{\pm0.68}$ & \underline{80.98}$_{\pm0.55}$ & 80.87$_{\pm0.29}$ & \underline{80.98}$_{\pm0.20}$ & 79.47$_{\pm0.86}$ & \textbf{81.38}$_{\pm0.60}$ \\
& DARNet & 81.35$_{\pm0.47}$ & \textbf{84.53}$_{\pm0.37}$ & 81.29$_{\pm0.30}$ & 81.08$_{\pm0.32}$ & \underline{81.36}$_{\pm0.34}$ & 81.01$_{\pm0.50}$ & 81.29$_{\pm0.39}$ & 81.31$_{\pm0.35}$ \\
\cmidrule{2-10}
& Average & 81.34 & 80.41 & 81.34 & 81.51 & \textbf{81.77} & \textbf{81.77} & 80.35 & \underline{81.73} \\
\bottomrule
\end{tabular}
\end{table*}

\subsubsection{Results on Generative Models}

To examine model-based brain signal generation, we evaluated 9 generative models across 3 generative paradigms and 3 generator architectures. Table~\ref{tab:gm_results} reports the average classification accuracies on two SSVEP datasets under the within-subject setting. Observe that:
\begin{itemize}
\item Overall, model-based generation improved the None baseline in most cases, but the gains were not uniform across classifiers and datasets. The improvement was modest on Nakanishi, where the best average accuracy increased from 89.63\% to 91.56\%. A larger gain was observed on the more challenging 40-target Benchmark dataset, where the best average accuracy increased from 71.07\% to 76.42\%.
\item Among the three generative paradigms, GAN-based models consistently achieved the best average performance on both datasets. In contrast, vanilla reconstruction models and VAE-based models provided only limited gains and even degraded strong classifiers in some cases, such as SSVEPNet on the Benchmark dataset. This indicated that downstream improvement is not guaranteed by using more complex generators. The advantage of GAN-based models may come from adversarial training, which encourages generated signals to be more informative for discrimination, whereas latent-regularized objectives may generate samples that are smooth but less useful for classification.
\item Generator architecture also influenced the results. Under the GAN-based setting, CNN-Transformer and CNN-LSTM generally performed better than CNN-only generators, suggesting that temporal modeling is useful for SSVEP generation. However, the effect of architecture was smaller than that of the generative paradigm, and the best architecture varied across classifiers. Therefore, model-based generation should be evaluated alongside the downstream decoder, rather than solely by the generator design.
\end{itemize}

\begin{table*}[htpb]
\centering
\small
\setlength{\tabcolsep}{0.35mm}
\renewcommand\arraystretch{1}
\caption{Average classification accuracies (\%) on two SSVEP datasets with four base classifiers under the within-subject setting, comparing nine generative models. The best average performance of each network is marked in bold, and the second best by an underline.}  \label{tab:gm_results}
\begin{tabular}{c|c|c|ccc|ccc|ccc}   \toprule
\multirow{2.5}{*}{Dataset} & \multirow{2.5}{*}{Backbone} & \multirow{2.5}{*}{None} & \multicolumn{3}{c|}{Vanilla} & \multicolumn{3}{c|}{GAN} & \multicolumn{3}{c}{VAE} \\
\cmidrule{4-12}
 &  &  & CNN & CNN-Trans. & CNN-LSTM & CNN & CNN-Trans. & CNN-LSTM & CNN & CNN-Trans. & CNN-LSTM \\
\midrule
\multirow{5.5}{*}{Nakanishi}
& FBCCA & 79.50$_{\pm4.0}$ & 79.44$_{\pm4.01}$ & 79.50$_{\pm4.2}$ & 79.78$_{\pm4.3}$ & \underline{86.06}$_{\pm4.2}$ & \textbf{86.33}$_{\pm3.0}$ & 85.94$_{\pm3.4}$ & 79.50$_{\pm4.1}$ & 79.61$_{\pm4.2}$ & 79.72$_{\pm4.4}$ \\
& EEGNet & 90.50$_{\pm3.9}$ & 87.56$_{\pm2.4}$ & \textbf{90.67}$_{\pm2.4}$ & 88.00$_{\pm2.2}$ & 89.22$_{\pm4.0}$ & 89.83$_{\pm1.7}$ & 90.39$_{\pm2.1}$ & 88.17$_{\pm2.9}$ & \underline{90.61}$_{\pm2.0}$ & 89.61$_{\pm1.2}$ \\
& CCNN & 91.33$_{\pm2.6}$ & 90.00$_{\pm4.5}$ & 91.17$_{\pm2.6}$ & 89.78$_{\pm2.7}$ & 92.50$_{\pm3.4}$ & \textbf{92.83}$_{\pm2.6}$ & \textbf{92.83}$_{\pm2.6}$ & 90.00$_{\pm4.5}$ & 91.11$_{\pm2.8}$ & 89.83$_{\pm2.6}$ \\
& SSVEPNet & 97.17$_{\pm0.7}$ & \textbf{97.33}$_{\pm1.3}$ & 96.61$_{\pm1.3}$ & 96.89$_{\pm0.6}$ & 97.06$_{\pm1.4}$ & \underline{97.22}$_{\pm1.5}$ & 97.06$_{\pm1.0}$ & \underline{97.22}$_{\pm1.4}$ & \underline{97.22}$_{\pm1.0}$ & 97.17$_{\pm1.3}$ \\
\cmidrule{2-12}
& Average & 89.63 & 88.58 & 89.49 & 88.61 & 91.21 & \underline{91.55} & \textbf{91.56} & 88.72 & 89.64 & 89.08 \\
\midrule
\multirow{5.5}{*}{Benchmark}
& FBCCA &63.08$_{\pm0.9}$ & 72.12$_{\pm0.8}$ & 72.20$_{\pm0.9}$ & 72.08$_{\pm0.9}$ & 77.68$_{\pm1.1}$ & \textbf{78.54}$_{\pm1.4}$ & \underline{77.96}$_{\pm1.0}$ & 72.19$_{\pm0.8}$ & 72.13$_{\pm0.9}$ & 72.08$_{\pm1.0}$ \\
& EEGNet & 61.72$_{\pm1.8}$ & 63.44$_{\pm0.9}$ & 62.90$_{\pm1.8}$ & 62.01$_{\pm2.0}$ & 66.53$_{\pm1.6}$ & \textbf{66.79}$_{\pm1.1}$ & \underline{66.72}$_{\pm2.0}$ & 63.96$_{\pm0.9}$ & 62.63$_{\pm1.8}$ & 61.73$_{\pm2.0}$ \\
& CCNN & 71.43$_{\pm1.1}$ & 68.94$_{\pm0.6}$ & 69.27$_{\pm1.4}$ & 69.33$_{\pm0.5}$ & 74.38$_{\pm1.3}$ & \underline{75.15}$_{\pm1.4}$ & \textbf{75.37}$_{\pm1.3}$ & 68.90$_{\pm0.7}$ & 69.27$_{\pm1.4}$ & 69.42$_{\pm0.5}$ \\
& SSVEPNet & \textbf{88.04}$_{\pm0.8}$ & 85.37$_{\pm0.9}$ & \underline{85.53}$_{\pm1.5}$ & 84.96$_{\pm1.4}$ & 84.92$_{\pm1.4}$ & 85.20$_{\pm1.1}$ & 85.29$_{\pm1.1}$ & 84.60$_{\pm0.8}$ & 85.30$_{\pm0.9}$ & 84.25$_{\pm1.4}$ \\
\cmidrule{2-12}
& Average & 71.07 & 72.47 & 72.48 & 72.10 & 75.88 & \textbf{76.42} & \underline{76.34} & 72.41 & 72.33 & 71.87 \\
\bottomrule
\end{tabular}
\end{table*}

\section{Synthetic Data Evaluation} \label{sect:sde}

The evaluation of generated data is pivotal for data-centric artificial intelligence. A scientifically grounded evaluation framework is essential for guiding both data generation and utilization \cite{long2024llms}. Giuffre \textit{et al.} \cite{giuffre2023harnessing} examined synthetic data applications in healthcare and highlighted concerns related to bias, quality, and privacy. Yan \textit{et al.} \cite{yan2022multifaceted} proposed a systematic framework for evaluating synthetic electronic health records from utility and privacy perspectives. Tosato \textit{et al.} \cite{tosato2023eeg} conducted qualitative and quantitative assessments of synthetic EEG data against real recordings. Long \textit{et al.} \cite{long2024llms} further evaluated synthetic data in terms of fidelity, diversity, and downstream task performance.

\begin{figure*}[htbp]\centering
\includegraphics[width=\linewidth,clip]{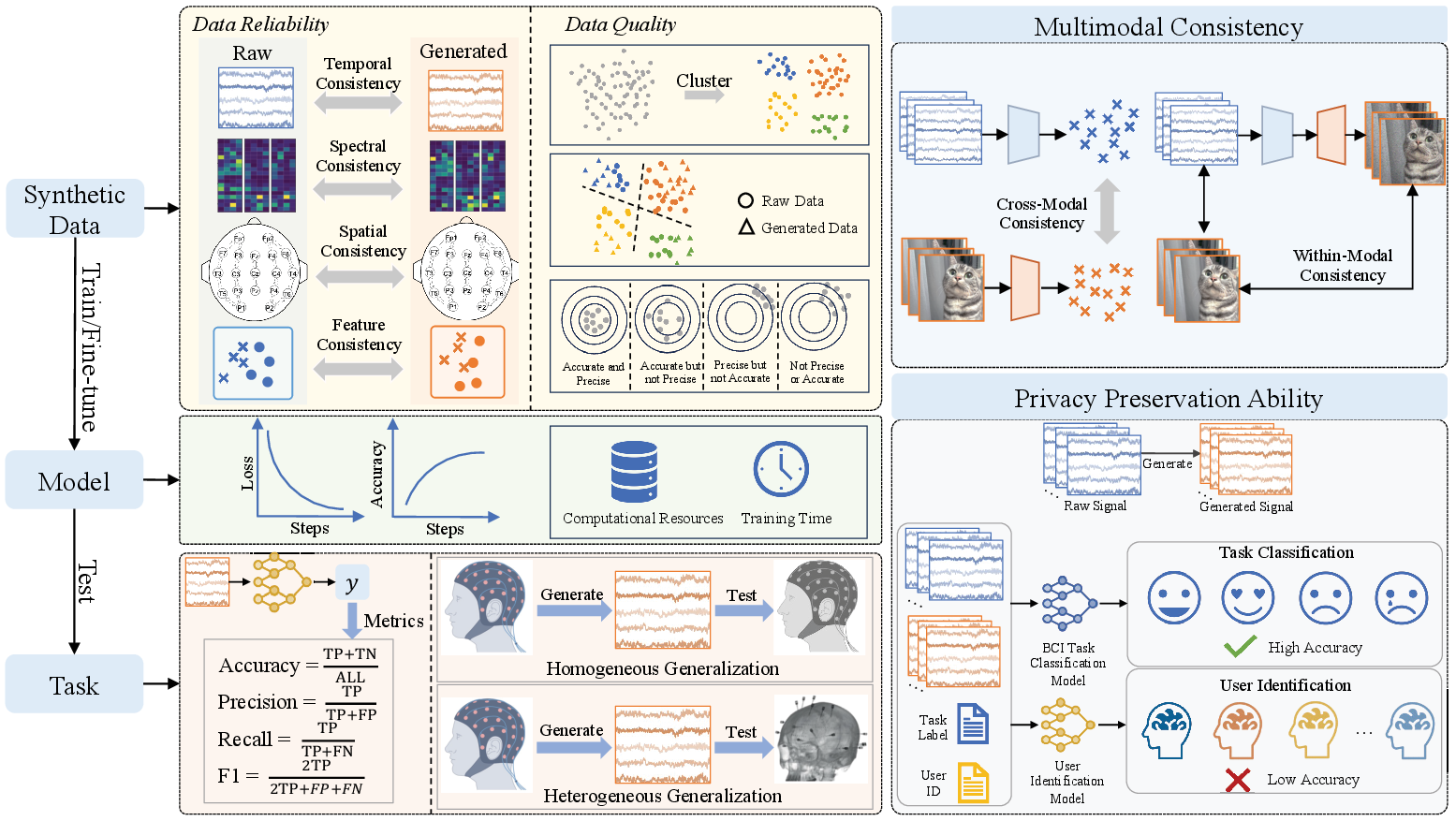}
\caption{Evaluation framework for generated data in BCIs, encompassing data reliability, quality, task performance, model performance, multimodal consistency, and privacy preservation ability.} \label{fig:evaluation}
\end{figure*}

For generated brain signals, evaluation should not be limited to whether synthetic data improve classification accuracy. A more rigorous assessment should distinguish four core questions: whether the generated signals resemble real recordings, whether they preserve physiologically meaningful neural patterns, whether they improve downstream BCI utility, and whether they avoid privacy leakage. Following this principle, we organize the evaluation framework into six dimensions: data reliability, data quality, task performance, model performance, multimodal consistency, and privacy preservation ability, as shown in Figure~\ref{fig:evaluation}. These dimensions jointly assess signal realism, physiological plausibility, downstream utility, and privacy risk.

\subsection{Data Reliability}

Data reliability evaluates whether generated signals preserve the essential characteristics of real brain recordings. It mainly reflects signal realism and physiological plausibility. Several aspects are commonly considered:
\begin{itemize}
\item \textit{Temporal Consistency}: Measures such as dynamic time warping, mutual information, and autocorrelation assess whether generated signals preserve temporal structures and repetitive patterns observed in real EEG signals.
\item \textit{Spectral Consistency}: Power spectral density and spectral similarity evaluate whether generated signals maintain realistic frequency-domain characteristics.
\item \textit{Spatial Consistency}: This examines whether generated data preserve correlations across channels and brain regions, reflecting plausible spatial organization of neural activity.
\item \textit{Feature Consistency}: Metrics such as Frechet inception distance (FID) \cite{Heusel2017FID}, maximum mean discrepancy, cosine similarity, and Jaccard similarity quantify similarity between generated and real data in feature space.
\item \textit{Signal Quality}: Signal quality can be assessed using metrics such as signal-to-noise ratio, which reflects the degree of interference or artifacts in generated signals.
\item \textit{Physiological Interpretability}: This assesses whether generated signals are consistent with known neural mechanisms, such as task-related spectral patterns, neural synchrony, and local-global brain activity consistency.
\end{itemize}

\subsection{Data Quality}

Data quality focuses on whether generated samples are useful and sufficiently diverse for model training. It complements reliability by examining the distributional coverage and informativeness of synthetic data.
\begin{itemize}
\item \textit{Diversity}: Measures whether generated samples cover the variability of the data distribution across subjects, cognitive states, and task conditions. Metrics such as the inception score (IS) \cite{salimans2016improved} and entropy-based statistics are commonly used.
\item \textit{Representativeness}: Assesses whether synthetic data reflect the target distribution of real signals, including their temporal, spectral, and spatial characteristics.
\item \textit{Uncertainty}: Reflects the model's confidence in generated samples. Incorporating uncertainty helps identify reliable and informative samples for training.
\end{itemize}

\subsection{Task and Model Performance}

Downstream utility is a central criterion for evaluating generated data in BCIs. It measures whether synthetic data can improve practical decoding performance, model robustness, and computational efficiency.
\begin{itemize}
\item \textit{Benchmark Performance}: Evaluates the effectiveness of generated data in downstream BCI tasks using metrics such as accuracy, balanced accuracy, recall, and F$_1$ score.
\item \textit{Generalization Ability}: Examines whether models trained with generated data generalize across subjects, sessions, datasets, devices, or modalities.
\item \textit{Training Stability}: Evaluates issues such as unstable optimization or mode collapse, particularly for adversarial generative models.
\item \textit{Training Efficiency}: Assesses computational cost, resource consumption, and training time required by generative models.
\end{itemize}

\subsection{Multimodal Consistency}

For translation-based generation, multimodal consistency is essential because generated brain signals should remain aligned with the conditioning modality. This dimension is particularly relevant when generation is performed across EEG, image, audio, text, or other physiological modalities.
\begin{itemize}
\item \textit{Within-Modal Consistency}: Evaluates similarity between generated signals and real data within the same modality in conditional latent space generation.
\item \textit{Cross-Modal Consistency}: Measures the alignment between latent representations of generated brain signals and other modalities in joint latent space generation.
\end{itemize}

\subsection{Privacy Preservation Ability}

Privacy preservation evaluates whether generated brain signals reduce the risk of exposing sensitive personal information while maintaining downstream utility. It should be noted that synthetic data are not inherently privacy-preserving, since generated samples may still leak identity-related or health-related information if the generator memorizes training data.
\begin{itemize}
\item \textit{Privacy Identification Protection}: Evaluates whether generated data can reveal user identity through privacy recognition models \cite{Chen2024user}. Lower identification accuracy indicates a reduced risk of identity leakage.
\item \textit{Privacy Inference Protection}: Assesses whether sensitive information can be reconstructed or inferred from generated data or model parameters through attacks such as gradient inversion or attribute prediction.
\end{itemize}

Overall, these evaluation dimensions are complementary. High downstream accuracy does not necessarily imply realistic or physiologically plausible generation, and realistic-looking signals may still carry privacy risks. Therefore, synthetic brain data should be evaluated from multiple perspectives, especially when they are used for clinical or privacy-sensitive BCI applications.

\section{Applications of Synthetic Data in BCIs}\label{sect:app}

\subsection{Enhancing Large Brain Model Training}

With the rapid development of large language models and vision-language models, large-scale brain models have emerged as a new frontier in BCI research. An increasing number of large-scale brain models have been proposed for universal feature extraction. EEGPT \cite{wang2025eegpt} is a pre-trained transformer model with 10 million parameters. LaBraM \cite{jianglarge2024} enables cross-dataset feature extraction with 5.8 million parameters. CBraMod \cite{wangcbramod2025} adopts a criss-cross transformer backbone to model spatial and temporal dependencies with 4.1 million parameters.

Beyond task-specific decoding paradigms, data generation is expected to play an increasingly important role in training large-scale brain models. These models aim to learn generalizable neural representations from large-scale EEG data across subjects, tasks, and recording conditions \cite{banville2021uncovering}. However, unlike text and vision domains with massive curated datasets, publicly available brain signals remain relatively limited, fragmented, and heterogeneous due to high acquisition costs, protocol variability, and privacy constraints \cite{roy2019deep}. In this context, data generation can expand the training data while preserving task-relevant information. Augmentation strategies based on signal processing and neurophysiological priors can also impose desirable invariances on learned representations.

Specifically, brain signal generation can support the entire lifecycle of large-scale brain model development, including both pre-training and downstream fine-tuning:
\begin{itemize}
\item \textit{Pre-Training}: When incorporated into pre-training objectives, these transformations can help models disentangle stable neural dynamics from interference factors such as noise, subject variability, and recording artifacts, improving robustness and cross-dataset transferability.
\item \textit{Fine-Tuning}: Synthetic data are particularly beneficial when labeled data are scarce. During the adaptation of large pre-trained models to downstream tasks, additional generated data can compensate for the limited task-specific samples and improve generalization.
\end{itemize}

\subsection{Improving Model Generalization Ability}

Homogeneous generalization ensures robust model performance across variations in data distributions, while heterogeneous generalization addresses the challenge of adapting to significantly different datasets or tasks, such as those arising from variations in data collection devices. Synthetic data facilitates both types of generalization by simulating diverse brain states and cognitive tasks, enriching training datasets with improved robustness. Integrating synthetic data enables models to adapt more effectively to both similar and highly diverse environments. Approaches like cross-subject data generation \cite{Wang2025CSDA} generate brain signals from both source and target subjects, reducing the gap for data generation under distribution shift.

\subsection{Privacy Preservation}

\subsubsection{Secure Data Sharing}

Sharing subjects' raw brain data across institutions may expose sensitive personal information. Generated brain signals provide a potential solution by preserving neural activity patterns while masking individual traits. This enables data aggregation from multiple sources to construct larger training sets without revealing private health information or mental states.

\subsubsection{Synthetic Data for Federated Learning}

Federated learning enables privacy-preserving collaborative training by keeping data locally and exchanging only model updates~\cite{Li2023Survey}. In general domains, data generation has been widely used to address statistical heterogeneity across clients~\cite{Huang2024Federated}. Existing studies employ strategies such as lightweight client-side augmentation to expand effective sample spaces~\cite{Zhang2023Data,Yan2025Simple,hao2021towards}, or server-side synthetic data to approximate global distributions and stabilize training~\cite{Chiaro2023FL,Peng2025Federated}.

However, the integration of data generation and federated learning in BCIs remains largely unexplored. Current work mainly focuses on client-side augmentation to improve cross-subject generalization~\cite{Liu2025mixEEG}, alleviate class imbalance~\cite{Lim2025FCEEG,Jia2026SAFE}, and enhance robustness through adversarial sample generation~\cite{Jia2026SAFE}. These strategies enable federated training without transferring sensitive data, thereby preserving privacy while improving model performance.

\subsection{Multimodal Data Generation}

\subsubsection{Cross-Modal Alignment}

Cross-modal alignment involves mapping data from one modality to another. By generating cross-modal signals, models can integrate information from multiple modalities, offering a more comprehensive understanding of brain activity.
\subsubsection{Cross-Modal Translation}

In BCIs, EEG-to-text or EEG-to-image translation could significantly improve communication for subjects with disabilities. For example, generating cross-modal data allows models to directly convert brain activity into text, speech, or visual outputs. This enables applications like brain-controlled typing, communication aids for non-verbal subjects, or real-time cognitive state monitoring.

\section{Discussion}\label{sect:discuss}

\subsection{Summary of Existing Literature}

Brain signal generation has evolved from simple signal transformations to sophisticated deep generative architectures. The choice of generation strategy involves a trade-off among downstream utility, signal fidelity, computational cost, and physiological interpretability.

Signal-transformation-based approaches are simple, efficient, and often interpretable, especially when they are guided by neurophysiological or signal-processing priors. However, their effectiveness depends strongly on whether the imposed transformation preserves task-relevant neural structures. Naive perturbations may improve robustness in some cases, but may also damage discriminative temporal, spectral, or spatial patterns. Model-based approaches, in contrast, learn data distributions directly and can generate more flexible samples, but they introduce new challenges such as black-box generation, limited biological plausibility, mode collapse in GANs, and sampling latency in diffusion models.

Feature-based generation remains a pragmatic strategy for addressing class imbalance and data scarcity, particularly when reliable handcrafted features are available. However, it may lose fine-grained temporal, spectral, and spatial information that is important for end-to-end decoding. Recent studies therefore show a growing interest in raw-signal generation, which better matches modern deep learning pipelines but also requires stricter evaluation of physiological plausibility and signal realism.

Translation-based generation frames brain signal generation as a cross-modal alignment problem. By leveraging auxiliary modalities such as images, audio, text, or other physiological signals, it provides a promising way to regularize sparse and noisy brain data. Future work should focus on robust cross-modal alignment, modality-specific uncertainty, and the distinction between brain-to-modality and modality-to-brain generation.

Overall, evaluation remains a central bottleneck. Standard metrics such as IS and FID may capture distributional similarity, but they do not necessarily reflect the functional relevance of generated brain signals. Therefore, future studies should jointly assess signal realism, physiological plausibility, downstream utility, and privacy leakage risk.

\subsection{When Does Synthetic Data Help BCI Decoding?}

Our benchmark results suggest that synthetic data are most useful when the generation process introduces meaningful variability while preserving task-relevant neural patterns. For MI, transformations that maintain temporal-spectral rhythms and spatial organization, such as DWTaug, provide more stable gains. For SSVEP, preserving frequency-locked periodicity and phase-related information is critical; approaches that disrupt phase synchronization, such as Flip, can severely degrade performance. For ESD, seizure-related waveforms are sensitive to unrealistic temporal or spatial perturbations, whereas spatially plausible transformations, such as CR, are more effective. For AAD, frequency domain perturbations and polarity-invariant transformations can be beneficial, but their effectiveness depends strongly on the downstream decoding model.

These findings indicate that performance gains do not always imply that generated signals are more realistic. Simple transformations may improve decoding by acting as regularization, increasing class diversity, or inducing useful invariances. Conversely, complex generative models may fail if they generate over-smoothed, poorly aligned, or physiologically implausible samples. Therefore, synthetic data should not be evaluated only by downstream accuracy, nor should more complex models be assumed to be better. The key question is whether the generated data preserve the task-relevant neural patterns while expanding the training distribution in a controlled way.

The choice between feature-based and raw-signal generation also depends on the application scenario. Feature-based generation is suitable when the discriminative representation is well understood or the dataset is imbalanced. Raw-signal generation is more appropriate when preserving temporal dynamics, spatial interactions, or frequency structures is important, especially for end-to-end models. Thus, the optimal generation strategy should be selected jointly with the BCI paradigm, the available data scale, and the downstream decoding models.

\subsection{Future Research Directions}

\subsubsection{Foundation Model Construction}

Large-scale synthetic datasets can support the training of EEG foundation models, particularly when real data are limited, fragmented, or difficult to share. Synthetic brain signals may be useful in both pre-training and fine-tuning by expanding neural variability and improving robustness across subjects, tasks, and recording conditions. However, future work should ensure that synthetic data improve generalizable representations rather than merely introducing task-specific shortcuts or artificial regularization.

\subsubsection{Federated BCIs}

Synthetic data can enhance federated BCI learning by balancing local data distributions and reducing cross-client heterogeneity \cite{hao2021towards}. This is particularly useful when institutions cannot directly share raw brain recordings. However, synthetic data are not inherently privacy-preserving, as generated samples may still leak subject-specific or health-related information. Future research should integrate privacy-aware generation with federated optimization and explicitly evaluate identity leakage and inference risks.

\subsubsection{Speech Decoding BCIs}

Speech decoding BCIs typically require long-term data collection to achieve reliable performance \cite{Willett2023}. To enlarge training sets under limited recording durations, existing studies have adopted signal-processing and neurophysiology-inspired augmentation strategies, including noise addition \cite{Jia2025,Mohan2025a}, wavelet transforms \cite{Mohan2025a}, sliding windows \cite{Balaji2017}, and spatial transformations \cite{Panachakel2021}. Generative models such as variational autoencoders have also been explored to learn compact and noise-resilient representations from speech-related neural signals \cite{Chen2025}. Given the large target vocabulary and limited neural data in speech decoding, data synthesis is expected to become an important direction for improving data efficiency and generalization.

\subsubsection{Medical Rehabilitation}

Rare neurological events, such as epileptic seizures or abnormal cognitive states, are often under-represented in brain datasets. Synthetic data generation can supplement these rare samples and improve the detection of clinically important events. However, medical applications require high physiological plausibility, interpretability, and safety. Future work should integrate domain constraints and explainable learning to ensure that generated signals correspond to meaningful neural activity rather than artifacts or spurious patterns.

\subsubsection{Real-Time BCIs}

Real-time BCIs require models to adapt to non-stationary brain states while maintaining low latency. Synthetic data can help improve robustness by simulating variations in brain activity and recording conditions. However, the computational cost of high-fidelity generation may limit real-time applicability. Lightweight generative models, online augmentation, model pruning, and efficient sampling strategies are therefore important directions for real-time BCI deployment.

\section{Conclusion}\label{sect:conclusions}

This paper presents a comprehensive survey of synthetic data generation for improving BCI model training, encompassing methodological taxonomies, benchmark evaluations, assessment metrics, and key applications. Brain signal generation has emerged as a promising strategy for producing diverse and useful training data for BCIs, while its signal realism, physiological plausibility, and privacy risks should be carefully evaluated. Existing approaches are systematically categorized into four classes: signal-transformation-based, feature-based, model-based, and translation-based generation. To enable fair comparison, we conducted benchmark experiments on representative generation approaches across four BCI paradigms and 11 public datasets, focusing on their downstream utility for BCI decoding. Further, we summarized evaluation principles for generated signals from multiple perspectives, including data reliability, signal quality, distribution consistency, decoding performance, and privacy preservation ability. By alleviating data scarcity, improving model generalization, and supporting the development of large-scale brain models, synthetic brain signals have the potential to advance more accurate, data-efficient, and privacy-aware BCI systems.

\section*{Code availability statement}
The code for this research is available at \url{https://github.com/wzwvv/DG4BCI}.

\section*{Acknowledgements}
This research was supported by the National Natural Science Foundation of China (625B2077 and 62525305).

\section*{Declarations}
The authors declare no competing interests.

\section*{Author Contributions}
Z.W. conceived the study, conducted the experiments, analysed the results, and wrote the manuscript. Z.H. and X.H. assisted with the experiments. H.W., T.J., J.L., S.L., and X.C. contributed to manuscript revision. D.W. supervised the study, provided funding support, and revised the manuscript. All authors reviewed and approved the final manuscript.

\bibliography{bci_dg}

\end{document}